\documentclass[twoside]{article}

\usepackage[accepted]{aistats2026}
\usepackage[round]{natbib}

\usepackage{amsmath}
\usepackage{amsthm}
\usepackage{amssymb}
\usepackage{mathtools}

\theoremstyle{plain}
\newtheorem{theorem}{Theorem}[section]

\theoremstyle{definition}

\theoremstyle{remark}

\usepackage{graphicx}
\usepackage{subfig}

\usepackage{booktabs}
\usepackage{multirow}
\usepackage{makecell}

\usepackage{optidef}

\usepackage{algorithm}
\usepackage{algpseudocode}

\usepackage{xurl}

\usepackage{hyperref}

\usepackage{siunitx}

\begin{document}

%

%

\twocolumn[

\aistatstitle{Power Transform Revisited: Numerically Stable, and Federated}

\aistatsauthor{Xuefeng Xu \And Graham Cormode}

\aistatsaddress{University of Warwick \And University of Oxford} ]

\begin{abstract}
Power transforms are popular parametric methods for making data more Gaussian-like, and are widely used as preprocessing steps in statistical analysis and machine learning. However, we find that direct implementations of power transforms suffer from severe numerical instabilities, which can lead to incorrect results or even crashes. In this paper, we provide a comprehensive analysis of the sources of these instabilities and propose effective remedies. We further extend power transforms to the federated learning setting, addressing both numerical and distributional challenges that arise in this context. Experiments on real-world datasets demonstrate that our methods are both effective and robust, substantially improving stability compared to existing approaches.
\end{abstract}

\section{Introduction}
\label{sec:intro}
Power transforms are widely used to stabilize variance and reduce skewness, and are described in textbooks as foundational tools for data preprocessing. Among them, the Box-Cox \citep{Box1964} and Yeo-Johnson \citep{Yeo2000} transforms are the most prominent, and are frequently employed in statistical analysis and machine learning pipelines \citep{Ruppert2001,Fink2009}. Despite their popularity, directly implementing these mathematical functions leads to serious numerical instabilities, producing erroneous outputs or even causing program crashes. This is not a theoretical concern: as we show in Table~\ref{tab:adv-power}, simple inputs can easily provoke such failures.

The issue has been identified in prior work such as the MASS package \citep{Venables2002}, but only partial solutions have been proposed. Recently, this issue was raised by \citet{Marchand2022}. Unfortunately, their analysis includes some incorrect claims and unsound solutions. As shown in Figure~\ref{fig:boxcox-expsearch}, their method fails to identify optimal parameters and can crash even on simple datasets. A very recent study by \citet{Barron2025} also mentions numerical issues, but their remedy relies on simple replacement of the numerical function, which remains insufficient to ensure stability. Consequently, this foundational question has been unanswered until now.

In this paper, we provide a comprehensive analysis of numerical instabilities in power transforms. Our solution includes log-domain computation, careful reformulation of critical expressions, and constraints on extreme parameter values, all of which are essential to ensure robust numerical behavior. Guided by our theoretical analysis, we also construct adversarial datasets that systematically trigger overflows under different floating-point precisions. Such data can occur naturally in real-world settings (e.g., Figure~\ref{fig:feature-hist} in the Appendix) or be deliberately introduced by adversarial clients in federated learning scenarios.

We further extend power transforms to the federated setting, where distributed data introduces unique challenges. We design a numerically stable one-pass variance aggregation method for computing the negative log-likelihood (NLL), enabling robust and efficient optimization of transformation parameters across multiple clients with minimal communication overhead.

Extensive experiments on real-world datasets validate the effectiveness of our approach. Compared to existing implementations, including those of \citet{Marchand2022}, our methods consistently achieve superior stability in both centralized and federated settings.

\textbf{Our contributions} are:
\begin{enumerate}
    \item We conduct a comprehensive analysis of numerical instabilities in power transforms, together with recipes for constructing adversarial datasets that expose these weaknesses.
    \item We propose remedies addressing each source of instability, yielding numerically stable algorithms in both centralized\footnote{Our implementation has been merged into \href{https://scipy.org/}{\textit{SciPy}}.} and federated learning settings.
    \item We perform extensive experimental validation on real-world datasets, demonstrating the effectiveness and robustness of our methods.
\end{enumerate}
The rest of the paper is organized as follows. Section~\ref{sec:prelim} reviews preliminaries. Section~\ref{sec:understand-instable} explains the root causes of instability, while Section~\ref{sec:stable-power} presents our remedies. Section~\ref{sec:fedpower} extends the approach to federated learning. Experimental results are presented in Section~\ref{sec:expt}, further issues are discussed in Section~\ref{sec:discussion}, and Section~\ref{sec:conclusion} concludes.

\section{Preliminaries}
\label{sec:prelim}
\subsection{Power Transform}
\label{sec:power-trans}
\begin{figure}[t]
\centering
\includegraphics[width=0.445\columnwidth]{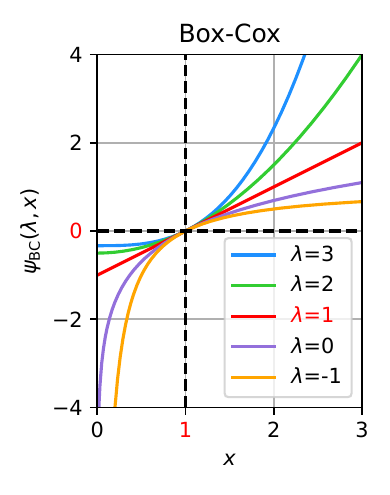}%
\label{fig:boxcox}%
\includegraphics[width=0.55\columnwidth]{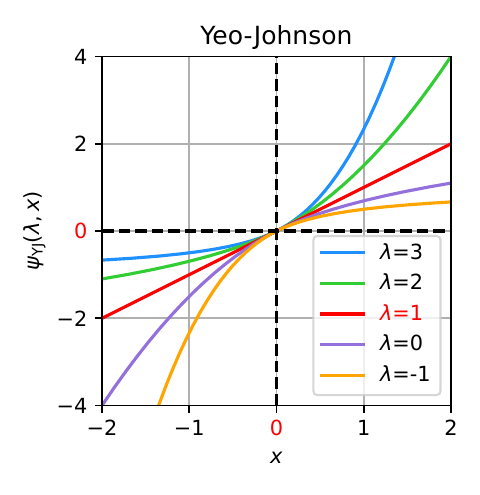}%
\label{fig:yeojohnson}%
\caption{Transformation functions for different $\lambda$.}%
\label{fig:power}%
\end{figure}

\begin{figure}[t]
\centering
\includegraphics[width=\columnwidth]{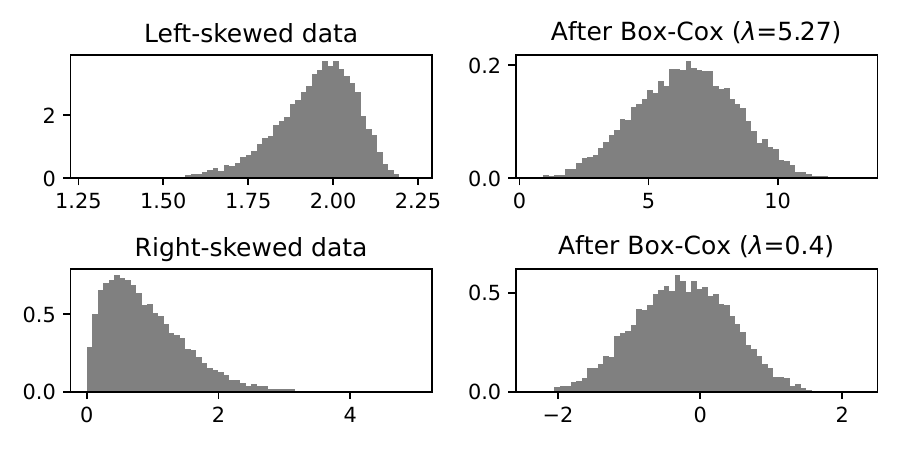}%
\caption{Box-Cox transformation on left- (top) and right-skewed (bottom) data.}%
\label{fig:boxcox-skew}%
\end{figure}

The Box-Cox (BC) transformation \citep{Box1964} is a widely used power transform requiring strictly positive inputs ($x > 0$):
\begin{equation}
\label{eq:boxcox}
\psi_{\text{BC}}(\lambda, x) =
\begin{cases}
\frac{x^\lambda-1}{\lambda} & \text{if } \lambda\neq0,\\
\ln x & \text{if } \lambda=0.
\end{cases}
\end{equation}
The Yeo-Johnson (YJ) transformation \citep{Yeo2000} extends BC to accommodate both positive and negative values:
\begin{equation}
\label{eq:yeojohnson}
\psi_{\text{YJ}}(\lambda, x) =
\begin{cases}
\frac{(x+1)^\lambda-1}{\lambda} & \text{if } \lambda\neq0,x\ge0,\\
\ln(x+1) & \text{if } \lambda=0,x\ge0,\\
\frac{(-x+1)^{2-\lambda}-1}{\lambda-2} & \text{if } \lambda\neq2,x<0,\\
-\ln(-x+1) & \text{if } \lambda=2,x<0.\\
\end{cases}
\end{equation}
The parameter $\lambda$ controls the shape of the transformation; when $\lambda \neq 1$, the mapping is nonlinear and alters the distribution. Figure~\ref{fig:power} shows the transformation curves for various $\lambda$, while Figure~\ref{fig:boxcox-skew} illustrates Box-Cox's effect on left- and right-skewed data.

The parameter $\lambda$ is typically estimated by minimizing the negative log-likelihood (NLL):
\begin{align}
\label{eq:nll-boxcox}
\text{NLL}_\text{BC} &= (1-\lambda) \sum_{i=1}^n \ln x_i
+ \frac{n}{2} \ln \sigma^2_{\psi_{\text{BC}}}, \\
\label{eq:nll-yeojohnson}
\text{NLL}_\text{YJ} &= (1-\lambda) \sum_{i=1}^n \text{sgn}(x_i) \ln(|x_i|+1)
+ \frac{n}{2} \ln \sigma^2_{\psi_{\text{YJ}}},
\end{align}
where $\sigma^2_{\psi} = \mathrm{Var}[\psi(\lambda, x)]$. Both NLL functions are strictly convex \citep{Kouider1995, Marchand2022}, guaranteeing a unique optimum. \textit{SciPy} \citep{Virtanen2020} estimates $\lambda^*$ using Brent's method \citep{Brent2013}, a derivative-free optimizer with superlinear convergence.
\subsection{Numerically Stable Variance Computation}
\label{sec:stable-var}
Variance is computed as $\frac{1}{n}\sum_{i=1}^n (x_i - \bar{x})^2$, where $\bar{x}$ is the sample mean. This requires two passes over the data, which is inefficient for large datasets or streaming data. The equivalent one-pass form,
\begin{equation}
\frac{1}{n} \sum_{i=1}^n x_i^2 - \frac{1}{n^2} \left(\sum_{i=1}^n x_i\right)^2,
\label{eq:unstable-var}
\end{equation}
is well known to be numerically unstable \citep{Ling1974, Chan1983}.

\begin{figure}[t]
\centering
\includegraphics[width=0.4\columnwidth]{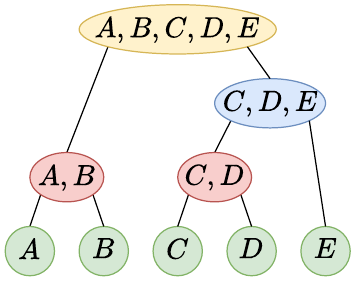}%
\caption{Tree-structured variance aggregation.}%
\label{fig:tree-aggregation}%
\end{figure}

A numerically stable one-pass approach \citep{Chan1982} maintains $(n, \bar{x}, s)$ triples, where $s = \sum_{i=1}^n (x_i - \bar{x})^2$. Merging two partial aggregates $(n^{(A)}, \bar{x}^{(A)}, s^{(A)})$ and $(n^{(B)}, \bar{x}^{(B)}, s^{(B)})$ is done as:
\begin{align}
n^{(AB)} &= n^{(A)} + n^{(B)}, \\
\bar{x}^{(AB)} &= \bar{x}^{(A)} + (\bar{x}^{(B)} - \bar{x}^{(A)}) \cdot \frac{n^{(B)}}{n^{(AB)}}, \label{eq:pairwise-mean} \\
s^{(AB)} &= s^{(A)} + s^{(B)} + (\bar{x}^{(B)} - \bar{x}^{(A)})^2 \cdot \frac{n^{(A)} n^{(B)}}{n^{(AB)}}. \label{eq:pairwise-ssd}
\end{align}
This method can be applied sequentially, but a tree-structured aggregation (Figure~\ref{fig:tree-aggregation}) minimizes error accumulation and improves numerical stability.

\subsection{Federated Learning}
\label{sec:fed-learning}
Federated learning \citep{Kairouz2021} is a distributed paradigm where multiple clients collaboratively train a model without sharing raw data, thereby preserving privacy. Each client computes local updates from its own data and transmits them to a central server, which aggregates the information to update the global model. Communication efficiency is a key concern, as excessive exchanges can be costly; methods such as FedAvg \citep{McMahan2017} are commonly used to reduce communication overhead.

\section{Understanding Numerical Instabilities}
\label{sec:understand-instable}
\begin{table*}[t]
\centering
\caption{Adversarial datasets causing negative or positive overflow under different floating-point precisions.}
\label{tab:adv-power}
\begin{tabular}{cccccccc}
\toprule
Transf. & Overflow & Adversarial Data & $\lambda^*$ & NLL & Extreme $\psi(\lambda^*,x)$ & Max Value \\
\midrule
\multirow{2}{*}{BC}
& Negative & [0.1, 0.1, 0.1, 0.109] & -41.70 & 23.91 & -1.20e+40 &
\multirow{4}{*}{\makecell{3.40e+38 \\ (Single)}} \\
& Positive & [10, 10, 10, 9.2] & 43.10 & 5.79 & 2.90e+41 & & \\
\multirow{2}{*}{YJ}
& Negative & [-10, -10, -10, -9.1] & -40.10 & 5.32 & -1.65e+42 & & \\
& Positive & [10, 10, 10, 9.1] & 42.10 & 5.32 & 1.65e+42 & & \\
\midrule
\multirow{2}{*}{BC}
& Negative & [0.1, 0.1, 0.1, 0.101] & -361.15 & 32.62 & -3.87e+358 &
\multirow{4}{*}{\makecell{1.80e+308 \\ (Double)}} \\
& Positive & [10, 10, 10, 9.9] & 357.55 & 14.18 & 9.96e+354 & & \\
\multirow{2}{*}{YJ}
& Negative & [-10, -10, -10, -9.9] & -391.49 & 14.18 & -1.51e+407 & & \\
& Positive & [10, 10, 10, 9.9] & 393.49 & 14.18 & 1.51e+407 & & \\
\midrule
\multirow{2}{*}{BC}
& Negative & [0.1, 0.1, 0.1, 0.10001] & -35936.9 & 51.03 & -2.30e+35932 &
\multirow{4}{*}{\makecell{1.19e+4932 \\ (Quadruple)}} \\
& Positive & [10, 10, 10, 9.999] & 35933.3 & 32.61 & 5.85e+35928 & & \\
\multirow{2}{*}{YJ}
& Negative & [-10, -10, -10, -9.999] & -39524.8 & 32.61 & -2.29e+41158 & & \\
& Positive & [10, 10, 10, 9.999] & 39526.8 & 32.61 & 2.292e+41158 & & \\
\midrule
\multirow{2}{*}{BC}
& Negative & [0.1, 0.1, 0.1, 0.100001] & -359353.0 & 60.24 & -2.74e+359347 &
\multirow{4}{*}{\makecell{1.61e+78913 \\ (Octuple)}} \\
& Positive & [10, 10, 10, 9.9999] & 359349.0 & 41.82 & 6.99e+359343 & & \\
\multirow{2}{*}{YJ}
& Negative & [-10, -10, -10, -9.9999] & -395283.0 & 41.82 & -6.47e+411640 & & \\
& Positive & [10, 10, 10, 9.9999] & 395285.0 & 41.82 & 6.47e+411640 & & \\
\bottomrule
\end{tabular}
\end{table*}
This section examines numerical instabilities in power transforms. We begin by explaining why such instabilities matter and how they can disrupt optimization. We then introduce adversarial datasets for testing implementation robustness.

\begin{figure}[t]
\centering
\includegraphics[width=0.5\columnwidth]{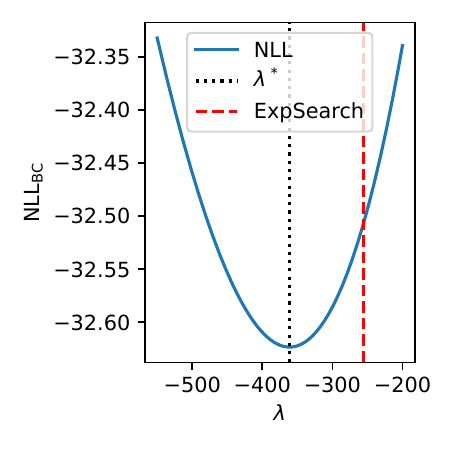}%
\label{fig:boxcox-expsearch-neg}%
\includegraphics[width=0.5\columnwidth]{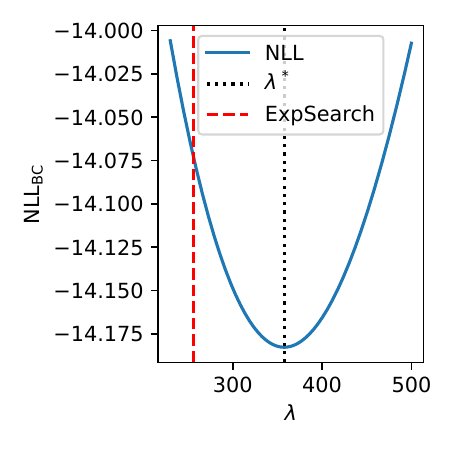}%
\label{fig:boxcox-expsearch-pos}%
\caption{%
    ExpSearch fails to find the true optimum $\lambda^*$.
    Left: data $[0.1, 0.1, 0.1, 0.101]$, $\lambda^*\approx -361$;
    Right: data $[10, 10, 10, 9.9]$, $\lambda^*\approx 358$.}%
\label{fig:boxcox-expsearch}%
\end{figure}

Numerical instabilities in power transforms arise in the computation of the NLL functions defined in Equations~\eqref{eq:nll-boxcox} and \eqref{eq:nll-yeojohnson}. These instabilities can disrupt optimization, producing suboptimal or even incorrect results. In practice, this means that the output of the method becomes unusable, causing bugs or failures in downstream tasks. Moreover, these issues cannot be resolved by simple quick fixes, as the instability is fundamental to the underlying computation and requires careful redesign for robust operation.  
As illustrated in Figure~\ref{fig:boxcox-expsearch}, the Exponential Search method \citep{Marchand2022}\footnote{There is a mismatch between the sign formula (Equation~3) and the \texttt{ExpUpdate} method (Algorithm~2) in \citep{Marchand2022}. The sign formula is $\partial\text{NLL}_{\text{YJ}} / \partial \lambda$ (not $\partial\ln\mathcal{L}_{\text{YJ}} / \partial\lambda$); therefore, $\Delta = 1$ in \texttt{ExpUpdate} should instead be $\Delta = -1$. Alternatively, one could add a negative sign in Equation~3 and keep \texttt{ExpUpdate} unchanged.} fails even on simple datasets. The true optimum $\lambda^*$ can be quite large in magnitude, causing numerical overflow when evaluating the power functions, e.g., $0.1^{-361}$ or $10^{358}$. As a result, ExpSearch will crash and return values that diverge significantly from the true optimum.

To better understand these issues, we first summarize key properties of the BC transformation $\psi(\lambda, x)$ in Theorem~\ref{thm:boxcox}; the proof is intricate and deferred to Appendix~\ref{proof:boxcox}. Analogous results for YJ can be found in \citet{Yeo1997,Yeo2000}.
\begin{theorem}
\label{thm:boxcox}
The Box-Cox transformation $\psi(\lambda,x)$ defined in \eqref{eq:boxcox} has the following properties:
\begin{enumerate}
\item $\psi(\lambda,x)\ge0$ for $x\ge1$, and $\psi(\lambda,x)<0$ for $x<1$. \label{thm:zero}
\item $\psi(\lambda,x)$ is convex in $x$ for $\lambda>1$ and concave in $x$ for $\lambda<1$. \label{thm:shape-x}
\item $\psi(\lambda,x)$ is a continuous function of $(\lambda,x)$. \label{thm:continuous}
\item If $\psi^{(k)}=\partial^k\psi(\lambda,x)/\partial\lambda^k$ then, for $k\ge1$:
\[ \psi^{(k)} =
\begin{cases}
    [x^\lambda(\ln x)^k-k\psi^{(k-1)}]/\lambda & \text{if } \lambda\neq0,\\
    (\ln x)^{k+1}/(k+1) & \text{if } \lambda=0.
\end{cases} \]
$\psi^{(k)}$ is continuous in $(\lambda,x)$ and $\psi^{(0)}\equiv\psi(\lambda,x)$. \label{thm:derivative}
\item $\psi(\lambda,x)$ is increasing in both $\lambda$ and $x$. \label{thm:increasing}
\item $\psi(\lambda,x)$ is convex in $\lambda$ for $x>1$ and concave in $\lambda$ for $x<1$. \label{thm:shape-lmb}
\end{enumerate}
\end{theorem}
These properties help pinpoint instability sources and guide the construction of adversarial datasets that trigger numerical overflow (Table~\ref{tab:adv-power}) under various floating-point precisions.  
Section~\ref{sec:stable-power} presents a numerically stable formulation to address these issues.  
In summary, there is a simple recipe that can lead to numerical overflow:

\textbf{Staying on One Side of Zero Points.}
By Theorem~\ref{thm:boxcox}.\ref{thm:zero}, $\psi(\lambda,1) = 0$ for all $\lambda$. Thus, $x=1$ should be avoided. Choosing all $x > 1$ ensures $\psi(\lambda,x) > 0$ and may lead to \emph{positive} overflow; similarly, choosing all $x < 1$ can yield \emph{negative} overflow. Datasets containing both $x > 1$ and $x < 1$ may be less likely to trigger overflow, and constructing simple adversarial examples in this regime is more challenging.

\textbf{Extreme Skewness.}
For $\lambda > 1$, $\psi(\lambda,x)$ is convex and increasing in $x$ (Theorem~\ref{thm:boxcox}.\ref{thm:shape-x} and \ref{thm:boxcox}.\ref{thm:increasing}), stretching the right tail more than the left.  
Thus, left-skewed data tends to push $\lambda > 1$ toward positive overflow after transformation; conversely, right-skewed data tends to push $\lambda < 1$ toward negative overflow. Extreme skewness drives $\lambda$ far from 1.

\textbf{Small Variance.}
Tightly clustered data also drives $\lambda$ to extreme values, as the transformation must expand interpoint distances to approach normality. For $x > 1$, $\psi(\lambda,x)$ is convex and increasing in $\lambda$ (Theorem~\ref{thm:boxcox}.\ref{thm:increasing} and \ref{thm:boxcox}.\ref{thm:shape-lmb}), so large $\lambda$ favors positive overflow; for $x < 1$, it is concave and increasing, so small $\lambda$ favors negative overflow. 
Combining extreme skewness with small variance (achieved by inserting duplicate values) is particularly effective at producing overflow.

As we show experimentally in Section~\ref{sec:expt}, such conditions naturally occur in real datasets or can be deliberately created by adversarial clients to poison training.
\section{Numerically Stable Power Transform}
\label{sec:stable-power}
This section presents a numerically stable approach for power transforms, addressing different sources of instability and introducing corresponding remedies.

Modern computers use finite-precision floating-point arithmetic (e.g., IEEE 754 standard \citep{IEEE-754-2019}), which limits the range of representable values. As shown in Table~\ref{tab:adv-power}, power transforms can generate extreme values that exceed these limits, resulting in numerical overflow. Therefore, direct computation of the NLL functions is prone to instability.

Beyond the representation issue, the choice of optimization method also plays a critical role. In particular, derivative-based methods, such as Exponential Search \citep{Marchand2022}, are prone to instability, whereas derivative-free methods (e.g., Brent's method) can achieve stable optimization. We detail these observations below.

\textbf{Derivative-based methods are unstable.}
Consider Exponential Search, which requires the first-order derivative of the NLL function. For the Box-Cox transformation, this derivative is
\begin{multline}
\frac{\partial\text{NLL}_\text{BC}}{\partial\lambda}
=\frac{1}{\sigma^2_{\psi_\text{BC}}}\Biggl(\sum_{i=1}^n\psi_\text{BC}(\lambda,x_i)\cdot\psi^{(1)}_\text{BC}(\lambda,x_i)\\
-\frac{1}{n}\sum_{i=1}^n\psi_\text{BC}(\lambda,x_i)\cdot\sum_{i=1}^n\psi^{(1)}_\text{BC}(\lambda,x_i)\Biggr)-\sum_{i=1}^n\ln x_i
\end{multline}
Here, terms such as $\psi_\text{BC}(\lambda,x)$, $\psi^{(1)}_\text{BC}(\lambda,x)$ (Theorem~\ref{thm:boxcox}.\ref{thm:derivative}), and $\sigma^2_{\psi_\text{BC}}$ can easily overflow and cannot be computed reliably. Consequently, derivative-based optimization is not numerically stable.

\textbf{Derivative-free methods are stable.}
By contrast, derivative-free methods rely only on evaluating the NLL function itself, which can be stabilized. In particular, the NLL function involves only $\ln\sigma^2_\psi$, which can be represented in floating-point format and efficiently computed using log-domain methods \citep{logvar} (see Appendix~\ref{sec:log-domain}).

\begin{figure}[t]
\centering
\subfloat[Eq. \eqref{eq:logvar-origin} vs. \eqref{eq:logvar-rm-const}]{%
\includegraphics[width=0.5\columnwidth]{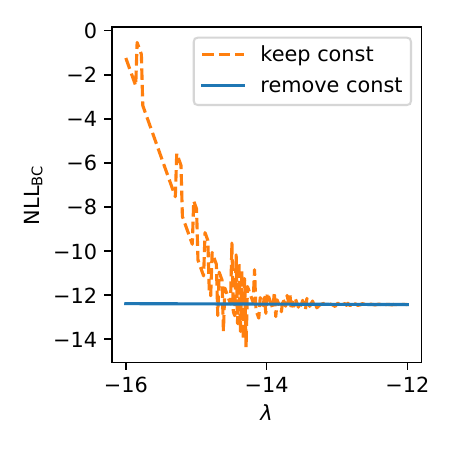}}%
\subfloat[Eq. \eqref{eq:logvar-rm-const} vs. \eqref{eq:logvar-lmd-out}]{%
\includegraphics[width=0.5\columnwidth]{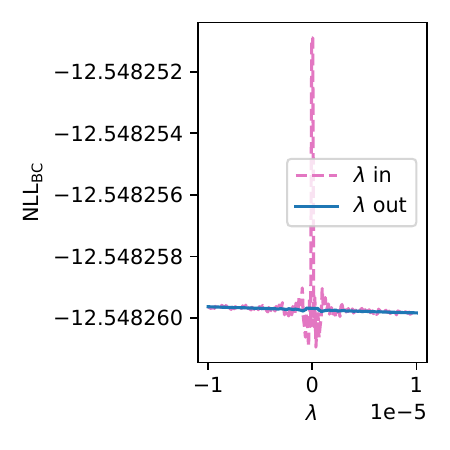}}%
\caption{%
    Comparison of NLL curves using different equations. Data used: [10, 10, 10, 9.9].}%
\label{fig:nll-stable}
\end{figure}

\textbf{Modifying the Variance Computation.}
While log-domain computation mitigates overflow, additional instabilities remain. Specifically, the computation of $\ln\sigma^2_{\psi_{\text{BC}}}$ requires careful reformulation:
\begin{align}
\ln\sigma^2_{\psi_{\text{BC}}} 
&= \ln\text{Var}[(x^\lambda-1)/\lambda] \label{eq:logvar-origin} \\
&= \ln\text{Var}(x^\lambda/\lambda) \label{eq:logvar-rm-const} \\ 
&= \ln\text{Var}(x^\lambda) -2\ln|\lambda| \label{eq:logvar-lmd-out}
\end{align}
Equation~\eqref{eq:logvar-rm-const} removes the constant $-1/\lambda$, which avoids catastrophic cancellation \citep{removeconst}. For example, with $\lambda<-14$ and $x>1$, we have $x^\lambda\to0$, and the subtraction $x^\lambda-1$ leads to severe precision loss. Therefore, computing the variance after the transformation becomes very unstable. The same occurs for large $\lambda$ with $x<1$. This instability, illustrated in the left panel of Figure~\ref{fig:nll-stable}, shows large fluctuations in the NLL curve when keeping the constant term, which will disrupt optimization whenever the evaluated $\lambda$ lies in such regions.

Further, as $\lambda\to0$, computing $x^\lambda/\lambda$ yields extreme values, destabilizing the variance computation. Factoring out $\lambda$, as in \eqref{eq:logvar-lmd-out}, restores stability. The right panel of Figure~\ref{fig:nll-stable} demonstrates this improvement, where the NLL curve becomes smooth after factoring out $\lambda$.

\textbf{Bounding extreme values.}
Finally, the optimal $\lambda^*$ may still yield extreme transformed values (Table~\ref{tab:adv-power}). To prevent this, we impose constraints during optimization. Since the transformation is monotone in both $\lambda$ and $x$ (Theorem~\ref{thm:boxcox}.\ref{thm:increasing}), we restrict the transformed data to lie within $[-y_B, y_B]$ by bounding $\lambda$ according to $x_{\text{max}}$ and $x_{\text{min}}$:
\begin{mini}|s|
	{\lambda}{\text{NLL}_{\text{BC}}(\lambda, x)}
	{}{}
	\addConstraint{\lambda\le\psi^{-1}_{\text{BC}}(x_{\text{max}},y_B)\ \ \ \text{if } x_{\text{max}}>1}
	\addConstraint{\lambda\ge\psi^{-1}_{\text{BC}}(x_{\text{min}},-y_B)\ \text{if } x_{\text{min}}<1}.
	\label{bc-constrained}
\end{mini}
Here, $\psi^{-1}_\text{BC}$ is the inverse Box-Cox transform, expressed via the Lambert $W$ function \citep{Corless1996}:
\begin{equation}
\psi^{-1}_{\text{BC}}(x,y)=
-\tfrac{1}{y}-\tfrac{1}{\ln x}W\left(-\tfrac{x^{-1/y}\ln x}{y}\right).
\label{eq:inv-boxcox}
\end{equation}
Since the Lambert $W$ function has two real branches ($k=0$ for $W(x)\ge-1$, and $k=-1$ for $W(x)\le-1$), we use the $k=-1$ branch in overflow cases ($x>1,\lambda\gg 1$ or $x<1,\lambda\ll 1$), where
\begin{align}
W\left(-\tfrac{x^{-1/y}\ln x}{y}\right)
&=-(\lambda+\tfrac{1}{y})\ln x \\
&\approx -\lambda\ln x = -\ln(x^\lambda) \ll -1 .
\end{align}
The same approach applies to the Yeo-Johnson transformation (see Appendix~\ref{sec:stable-yj}).
The default bound $y_B$ can be set to the maximum representable floating-point value to prevent overflow after transformation. Practitioners may select a smaller bound based on domain-specific requirements or constraints on downstream models.

\section{Federated Power Transform}
\label{sec:fedpower}
This section extends power transforms to the federated learning setting. We begin by explaining the challenges of federated NLL evaluation, then show how textbook variance computation can cause numerical instability in this context. Finally, we introduce a numerically stable variance aggregation method that enables reliable NLL evaluation under federation.

In federated learning, the objective is to find the global optimum $\lambda^*$ that minimizes the NLL across multiple clients, each holding its own local dataset. Standard federated optimization methods such as FedAvg do not apply here: each client may have a different local optimum $\lambda_j^*$, and averaging them does not in general recover the global optimum. For heterogeneous data distributions, the averaged value may diverge significantly from the true global solution. Consequently, the server must evaluate the NLL function on aggregated statistics and apply a derivative-free optimizer (e.g., Brent's method) to locate the global optimum.

\begin{figure}[t]
\centering
\includegraphics[width=0.6\columnwidth]{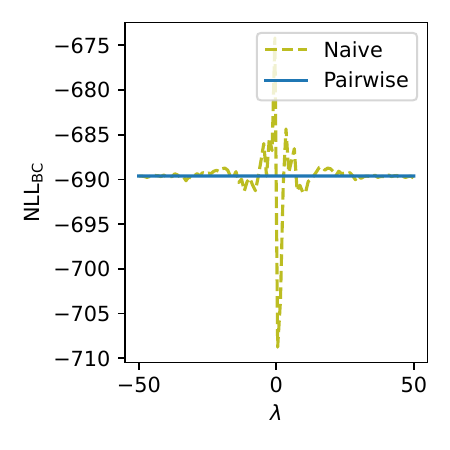}%
\caption{%
    Comparison of NLL curves using the naive and pairwise aggregation. Data are synthetic Gaussian samples from $\mathcal{N}(10^4, 10^{-3})$ with $100$ points.}%
\label{fig:fednll-onepass}%
\end{figure}

To reduce communication rounds, one-pass variance computation is preferred. However, this approach introduces severe numerical instabilities, which directly affect both NLL evaluation and the optimization of $\lambda$. To illustrate this, we compare two approaches: the naive one-pass method (Equation~\eqref{eq:unstable-var}, also used in \citet{Marchand2022}) and our numerically stable aggregation procedure (Algorithm~\ref{alg:var-agg}). As shown in Figure~\ref{fig:fednll-onepass}, the naive method produces large fluctuations in the NLL curve, disrupting optimization. In contrast, our stable approach yields smooth curves and enables reliable optimization.

For numerically stable federated NLL computation with Box-Cox, each client sends four values to the server: the local sum $c$, sample size $n$, mean $\bar{y}$, and sum of squared deviations $s$ for the transformed data (see line~\ref{alg:fednll-boxcox-client-send} in client part). The server aggregates these $(n, \bar{y}, s)$ triplets using the queue-based procedure in Algorithm~\ref{alg:var-agg} (line~\ref{alg:fednll-boxcox-server-agg} in server part), which ensures numerical stability compared to naive sequential aggregation.
\begin{algorithm}[t]
\caption{Variance Aggregation}
\label{alg:var-agg}
\begin{algorithmic}[1]
\State \textbf{Input:} Queue $Q$ containing $(n^{(j)}, \bar{y}^{(j)}, s^{(j)})$
\While{$|Q| > 1$}
    \State Dequeue $(n^{(A)}, \bar{y}^{(A)}, s^{(A)})$, $(n^{(B)}, \bar{y}^{(B)}, s^{(B)})$
    \State Compute $(n^{(AB)}, \bar{y}^{(AB)}, s^{(AB)})$ using \textbf{Pairwise}
    \State Enqueue $(n^{(AB)}, \bar{y}^{(AB)}, s^{(AB)})$ back into $Q$
\EndWhile
\State \Return the final $(N, \bar{Y}, S)$
\setcounter{ALG@line}{0}

\Statex
\Statex \textbf{Pairwise:}
\State \textbf{Input:} $(n^{(A)}, \bar{y}^{(A)}, s^{(A)})$ and $(n^{(B)}, \bar{y}^{(B)}, s^{(B)})$
\State $n^{(AB)} \gets n^{(A)} + n^{(B)}$
\State $\delta \gets \bar{y}^{(B)}-\bar{y}^{(A)}$
\State $\bar{y}^{(AB)} \gets \bar{y}^{(A)} + \delta\cdot\frac{n^{(B)}}{n^{(AB)}}$
\State $s^{(AB)} \gets s^{(A)} + s^{(B)} + \delta^2\cdot\frac{n^{(A)} n^{(B)}}{n^{(AB)}}$
\State \Return $(n^{(AB)}, \bar{y}^{(AB)}, s^{(AB)})$
\end{algorithmic}
\end{algorithm}
\begin{algorithm}[t]
\caption{Federated NLL Comptuation (Box-Cox)}
\label{alg:fed-nll-bc}
\begin{algorithmic}[1]
\Statex \textbf{Client Part:}
\State \textbf{Input:} $\lambda$ and local data $x_i$ (size $n$)
\State $c \gets \sum_i \ln x_i$
\State $y_i \gets
\begin{cases}
x_i^\lambda & \text{if } \lambda\neq0,\\
\ln x_i & \text{if } \lambda=0.
\end{cases}$ \label{alg:fednll-boxcox-client-rmconst}
\State $\bar{y} \gets \tfrac{1}{n} \sum_i y_i$ and $s \gets \sum_i (y_i-\bar{y})^2$
\State \textbf{Send:} $(c, n, \bar{y}, s)$ to server \label{alg:fednll-boxcox-client-send}
\setcounter{ALG@line}{0}

\Statex
\Statex \textbf{Server Part:}
\State \textbf{Input:} $\lambda$
\State Collect $(c^{(j)}, n^{(j)}, \bar{y}^{(j)}, s^{(j)})$ from clients
\State Enqueue $(n^{(j)}, \bar{y}^{(j)}, s^{(j)})$ into $Q$
\State Compute $(N, \bar{Y}, S)$ from $Q$ using Algorithm~\ref{alg:var-agg} \label{alg:fednll-boxcox-server-agg}
\If{$\lambda\neq0$}
\State $\ln S \gets \ln S - 2\ln|\lambda|$ \label{alg:fednll-boxcox-server-lmbout}
\EndIf
\State $\ln\sigma^2_{\psi} \gets \ln S - \ln N$
\State $\text{NLL}_\text{BC} \gets (1-\lambda)\sum_j c^{(j)} + \tfrac{N}{2}\ln\sigma^2_{\psi}$
\State \Return $\text{NLL}_\text{BC}$
\end{algorithmic}
\end{algorithm}

We also incorporate the numerically stable strategies from Section~\ref{sec:stable-power}, namely log-domain computation and modified variance formulations (see line \ref{alg:fednll-boxcox-client-rmconst} in client part and line \ref{alg:fednll-boxcox-server-lmbout} in server part). The case of Yeo-Johnson is more involved, since it handles both positive and negative inputs; details are deferred to Appendix~\ref{sec:fed-nll-yj}.
\section{Empirical Evaluation}
\label{sec:expt}
Our experiments contain two parts: (1) evaluating the effect of power transforms on downstream tasks, and (2) testing the numerical stability of our methods. Code is available at \url{https://github.com/xuefeng-xu/powertf}.
\subsection{Downstream Effectiveness}
\label{sec:downstream-expt}
\begin{figure*}[t]
\centering
\includegraphics[width=0.33\textwidth]{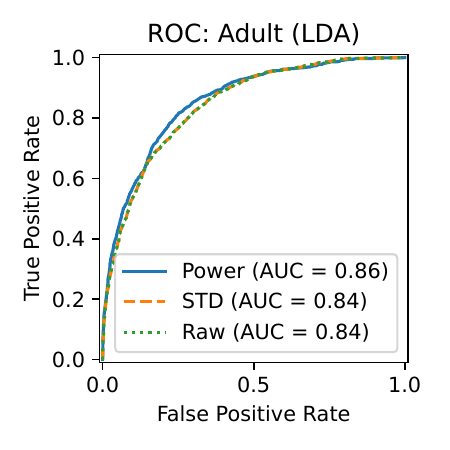}%
\includegraphics[width=0.33\textwidth]{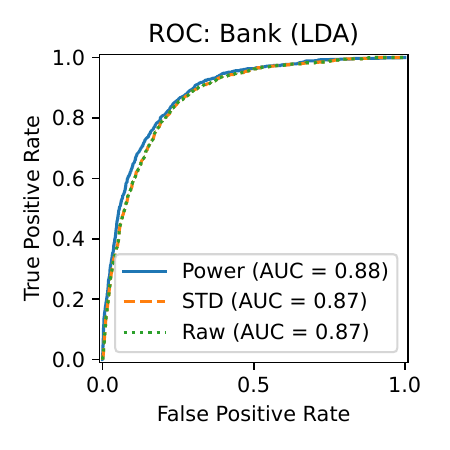}%
\includegraphics[width=0.33\textwidth]{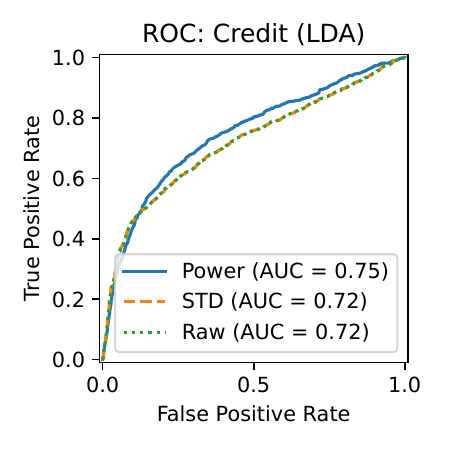}%
\caption{ROC curves after applying different transforms (LDA).}%
\label{fig:effect-roc-LDA}%
\end{figure*}
\begin{figure*}[t]
\centering
\includegraphics[width=0.33\textwidth]{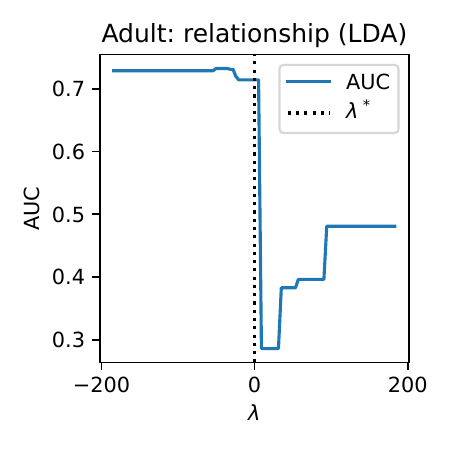}%
\includegraphics[width=0.33\textwidth]{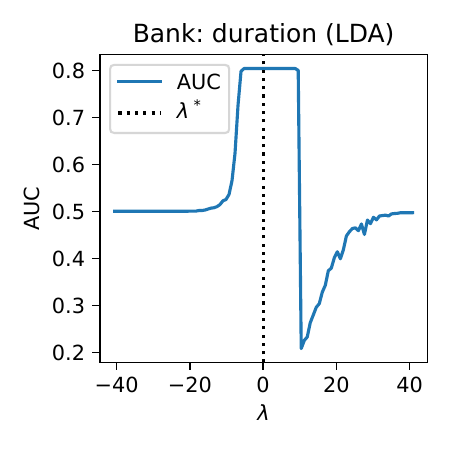}%
\includegraphics[width=0.33\textwidth]{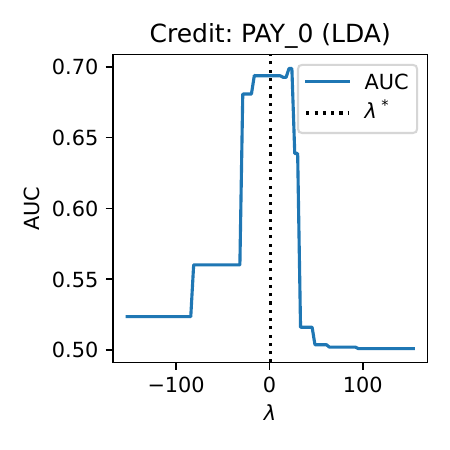}%
\caption{Effect of varying $\lambda$ on AUC scores (LDA).}%
\label{fig:deviate-auc-LDA}%
\end{figure*}
We first evaluate the impact of power transforms on downstream classification tasks using three datasets: Adult \citep{Becker1996}, Bank \citep{Moro2012}, and Credit \citep{Yeh2009}. Dataset statistics are shown in Table~\ref{tab:train-data-stats}. Each dataset is split into 80\% training and 20\% testing. Since features may contain negative values, we apply the Yeo-Johnson transformation to each feature, estimating $\lambda^*$ from the training set. We compare against two baselines:
(1) Standardization (STD: zero mean and unit variance), and (2) Raw data without any transformation.

We then train three classifiers on the transformed features: Linear Discriminant Analysis (LDA), Logistic Regression (LR), and XGBoost (XGB). Since LDA assumes normally distributed inputs, power transforms are expected to improve performance. Figure~\ref{fig:effect-roc-LDA} shows ROC curves and AUC scores on the test set (additional plots are in Figure~\ref{fig:effect-roc-LR-XGB} in the Appendix). Power transforms consistently outperform the baselines, although the improvements are modest. Notably, for XGBoost, the performance differential between using raw and preprocessed data is negligible. A recent study by \citet{Eftekhari2025} applied power transforms to hidden layers of deep neural networks, showing that improved Gaussianity leads to higher classification accuracy. This further confirms the effectiveness of power transforms.

We next study the effect of varying $\lambda$ on AUC scores. This tests whether the estimated $\lambda^*$ is indeed optimal for downstream tasks. Since each dataset has multiple features, we select the feature with the highest mutual information with the label and ignore other features and apply the Yeo-Johnson transform with varying $\lambda$. Figures~\ref{fig:deviate-auc-LDA} shows AUC as a function of $\lambda$ (additional results are in Figure~\ref{fig:deviate-auc-LR-XGB} in the Appendix).

Interestingly, $\lambda^*$ (marked with a vertical dashed line) does not always yield the absolute highest AUC (e.g., Adult dataset). However, it consistently provides competitive results across datasets, while deviating from it often reduces performance. This highlights the importance of numerically stable estimation of $\lambda^*$, as instability can significantly degrade downstream outcomes.
\subsection{Numerical Experiments}
\label{sec:numerical-expt}
\begin{table*}[t]
\caption{Dataset statistics and unstable features detected.}
\label{tab:numerical-data-stats}
\centering
\begin{tabular}{ccccccc}
\toprule
Datasets & \# Row & \# Col & \# Inst. & Transf. & ExpSearch & Linear \\
\midrule
\multirow{2}{*}{Blood} & \multirow{2}{*}{748} & \multirow{2}{*}{4} & \multirow{2}{*}{1}
& BC & --- & --- \\
& & & & YJ & --- & Monetary \\
\midrule
\multirow{2}{*}{Cancer} & \multirow{2}{*}{569} & \multirow{2}{*}{31} & \multirow{2}{*}{3}
& BC & --- & --- \\
& & & & YJ & --- & ID, area1, area3 \\
\midrule
\multirow{2}{*}{Ecoli} & \multirow{2}{*}{336} & \multirow{2}{*}{8} & \multirow{2}{*}{2}
& BC & --- & --- \\
& & & & YJ & lip, chg & lip, chg\\
\midrule
\multirow{2}{*}{House} & \multirow{2}{*}{1460} & \multirow{2}{*}{80} & \multirow{2}{*}{3}
& BC & YearRemodAdd, YrSold & YrSold \\
& & & & YJ & Street, YearRemodAdd, YrSold & YrSold \\
\bottomrule
\end{tabular}
\end{table*}
\begin{figure*}[t]
\centering
\includegraphics[width=\textwidth]{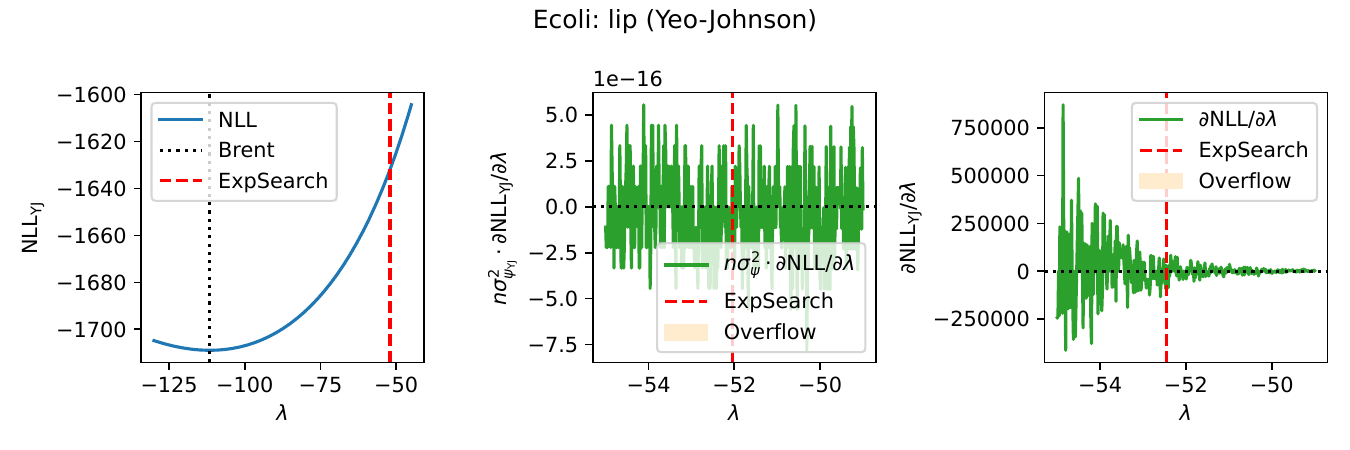}%
\caption{Comparison of Brent's method and ExpSearch.}%
\label{fig:brent-expsearch}%
\end{figure*}
\begin{figure*}[t]
\centering
\includegraphics[width=0.33\textwidth]{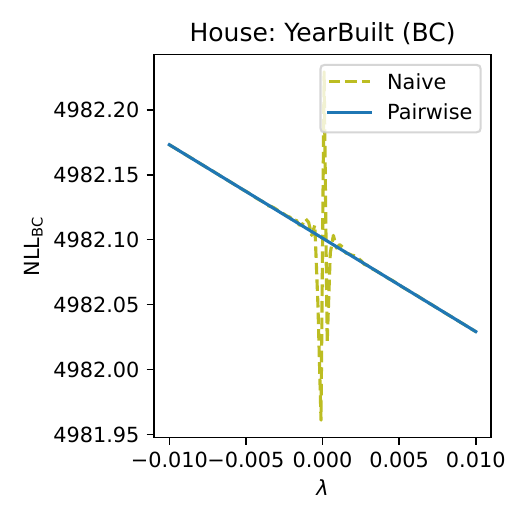}%
\includegraphics[width=0.33\textwidth]{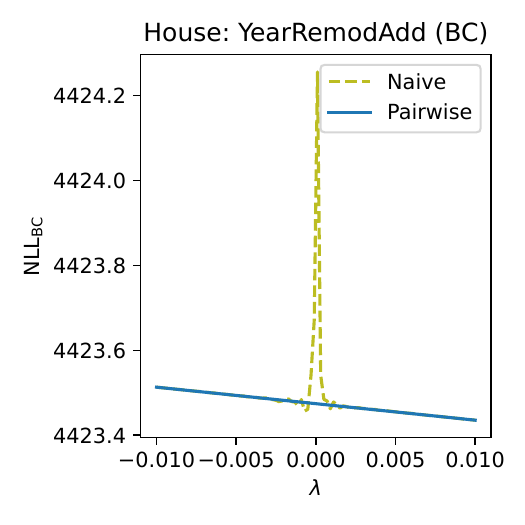}%
\includegraphics[width=0.33\textwidth]{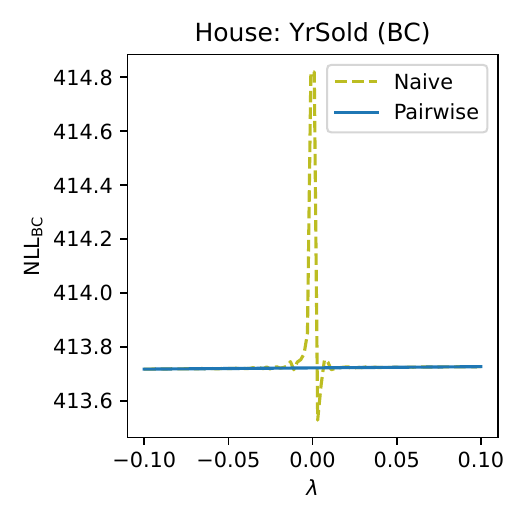}%
\caption{Comparison of pairwise and naive variance aggregation.}%
\label{fig:pairwise-naive}%
\end{figure*}
We now test numerical stability of power transforms on four datasets: Blood \citep{Yeh2008}, Cancer \citep{Wolberg1995}, Ecoli \citep{Nakai1996}, and House \citep{Montoya2016}. The first three were previously identified by \citet{Marchand2022} as unstable cases; House is a new addition with features exhibiting extreme skewness. We benchmark three methods:
(1) ExpSearch, a derivative-based optimization method \citep{Marchand2022} vs. our derivative-free approach (based on Brent's method);
(2) Linear-domain computation with our proposed log-domain method; and 
(3) Naive one-pass variance computation vs. our pairwise computation in federated learning.

Table~\ref{tab:numerical-data-stats} summarizes dataset statistics and nine features we identified with numerical issues. Histograms in Figure~\ref{fig:feature-hist} (in the Appendix) show these features are highly skewed, sometimes binary, and often extremely imbalanced. These match our analysis in Section~\ref{sec:understand-instable}, and align with our recipe for constructing adversarial datasets (Table~\ref{tab:adv-power}). Importantly, this shows that such pathological data naturally occur in practice, underscoring the need for stability in power transforms.

Next, we observe that seven features cause ExpSearch to fail. Instead of finding $\lambda^*$, it returns either a boundary of the search interval or an arbitrary value. Figures~\ref{fig:brent-expsearch} and \ref{fig:brent-expsearch-appendix-1} (in the Appendix) illustrate this. The left plot shows the true NLL curve and the optimum $\lambda^*$ (found by Brent's method). ExpSearch's $\lambda$ is far from optimal. The middle and right plots show the modified derivative used by ExpSearch versus the true derivative. Both are either unstable (returning arbitrary values) or overflow (returning boundary solutions). This explains ExpSearch's failures.

We then compare linear-domain versus log-domain computation. Eight features exhibit failures in the linear domain. Figure~\ref{fig:log-linear} (in the Appendix) shows that linear-domain NLL curves often overflow, preventing discovery of the optimum if $\lambda^*$ lies in the overflow region. By contrast, log-domain computation yields smooth and stable curves, reliably identifying $\lambda^*$. This confirms the robustness of our approach.

Finally, we evaluate federated NLL computation. We split each dataset into 100 clients and compare our variance aggregation method with naive one-pass variance computation. Figures~\ref{fig:pairwise-naive} and \ref{fig:pairwise-naive-appendix} (in the Appendix) show that our method produces smooth NLL curves, while the naive method introduces spikes that can disrupt optimization. This highlights the necessity of numerically stable aggregation in federated settings.
\section{Discussion}
\label{sec:discussion}
\textbf{Runtime performance} is an important consideration for practical deployment. Our stable recipe in Section~\ref{sec:stable-power} modifies the NLL computation and is indeed slower than the naive implementation due to the use of log-domain variance computation, as shown in Table~\ref{tab:time-compare} (Appendix). However, for scientific software, numerical correctness must take priority over raw speed. Moreover, the runtime remains well below human-perceptible latency in all cases, making it acceptable for most practical applications. In federated settings, our current implementation of variance aggregation (Algorithm~\ref{alg:var-agg}) processes clients sequentially. This can be further optimized by parallelizing the pairwise aggregation steps, thereby reducing overall latency without affecting numerical stability.

\textbf{Communication cost} is a common concern in federated learning, consisting of two components: (1) the size of each message per round and (2) the total number of communication rounds. In our method, each client only needs to send four numbers to the server, while the server sends back a single number, making the per-round communication negligible. For the number of rounds, we adopt Brent's method, which converges superlinearly. In practice, convergence typically requires 20-30 rounds, depending on the dataset, which is acceptable in most settings. One possible extension is to reduce the number of rounds by increasing message size, since the server can evaluate the NLL at multiple points per round. For example, after identifying a bracket that contains the minimum, a grid search could be applied. As shown in Figure~\ref{fig:comm-round-grid} (in the Appendix), communication rounds can be reduced to under 10 with a grid size of 1K. This trade-off between message size and number of rounds can be tuned for specific applications.

\textbf{Privacy} is a key concern in federated learning. Approaches such as Secure Aggregation \citep{Bonawitz2017}, Trusted Execution Environments (TEE) \citep{Sabt2015}, and Secure Multiparty Computation (SMPC) \citep{Lindell2020} can ensure that only aggregated statistics are revealed to the server, not individual client data. Our method uses pairwise variance aggregation for numerical stability. To integrate with SMPC, the division operations in Equations~\eqref{eq:pairwise-mean} and \eqref{eq:pairwise-ssd} can be handled by treating the sample count $n$ as a public value, which is standard in federated learning and also adopted by \citet{Marchand2022}. Preserving privacy during iterative optimization without revealing intermediate results in each round is a stronger requirement and we leave it for future work.

\textbf{Client sampling} is a common technique in federated learning to handle offline or dropout clients. When a random subset of clients is selected each round, the NLL estimate becomes biased. However, normalizing the NLL by the subsample size recovers unbiasedness without affecting $\lambda$ optimization. Specifically, the Box-Cox NLL $(1-\lambda)\sum_i \ln x_i + \frac{n}{2}\ln\sigma^2_{\psi_\text{BC}}$ divided by $n$ yields unbiased estimates of both the log-mean term and log-variance term under uniform random sampling. Thus, the normalized NLL can be reliably estimated even with random client subsampling.
\section{Conclusion}
\label{sec:conclusion}
Numerical issues have long been a challenge in scientific computing, as mathematically equivalent expressions can yield vastly different results under finite-precision arithmetic. In this paper, we addressed the numerical instability of power transforms, a widely used technique for data normalization. We conducted a detailed analysis of the sources of instability and proposed numerically stable approaches that combines log-domain computation, reformulated expressions, and bounding strategies. We further extended power transforms to the federated learning setting, introducing a numerically stable variance aggregation method suitable for distributed data. Our empirical results demonstrate the effectiveness and robustness of our methods in both centralized and federated scenarios. We believe that our work not only makes power transforms more reliable in practice, but also provides insights that can be applied to a broader range of numerical stability problems in scientific computing.
\section*{Acknowledgement}
We thank Matt Haberland for helpful discussions and feedback on our implementation in the GitHub pull request to \textit{SciPy}. We also thank the anonymous reviewers for their valuable comments.
XX is supported by a scholarship from the Department of Computer Science, University of Warwick. 
GC is supported in part by EPSRC grant EP/V044621/1.

\bibliography{powertf}

\clearpage
\section*{Checklist}

\begin{enumerate}

  \item For all models and algorithms presented, check if you include:
  \begin{enumerate}
    \item A clear description of the mathematical setting, assumptions, algorithm, and/or model. [Yes]
    \item An analysis of the properties and complexity (time, space, sample size) of any algorithm. [Yes]
    \item (Optional) Anonymized source code, with specification of all dependencies, including external libraries. [Yes]
  \end{enumerate}

  \item For any theoretical claim, check if you include:
  \begin{enumerate}
    \item Statements of the full set of assumptions of all theoretical results. [Yes]
    \item Complete proofs of all theoretical results. [Yes]
    \item Clear explanations of any assumptions. [Yes]     
  \end{enumerate}

  \item For all figures and tables that present empirical results, check if you include:
  \begin{enumerate}
    \item The code, data, and instructions needed to reproduce the main experimental results (either in the supplemental material or as a URL). [Yes]
    \item All the training details (e.g., data splits, hyperparameters, how they were chosen). [Yes]
    \item A clear definition of the specific measure or statistics and error bars (e.g., with respect to the random seed after running experiments multiple times). [Yes]
    \item A description of the computing infrastructure used. (e.g., type of GPUs, internal cluster, or cloud provider). [Yes]
  \end{enumerate}

  \item If you are using existing assets (e.g., code, data, models) or curating/releasing new assets, check if you include:
  \begin{enumerate}
    \item Citations of the creator If your work uses existing assets. [Not Applicable]
    \item The license information of the assets, if applicable. [Not Applicable]
    \item New assets either in the supplemental material or as a URL, if applicable. [Not Applicable]
    \item Information about consent from data providers/curators. [Not Applicable]
    \item Discussion of sensible content if applicable, e.g., personally identifiable information or offensive content. [Not Applicable]
  \end{enumerate}

  \item If you used crowdsourcing or conducted research with human subjects, check if you include:
  \begin{enumerate}
    \item The full text of instructions given to participants and screenshots. [Not Applicable]
    \item Descriptions of potential participant risks, with links to Institutional Review Board (IRB) approvals if applicable. [Not Applicable]
    \item The estimated hourly wage paid to participants and the total amount spent on participant compensation. [Not Applicable]
  \end{enumerate}

\end{enumerate}

\clearpage
\appendix
\onecolumn
\section{Proof of Theorem~\ref{thm:boxcox}}
\label{proof:boxcox}
\begin{enumerate}
\item
For $x\ge1$, we have:
\begin{equation}
    \begin{cases}
        x^\lambda-1 \ge 0 & \text{if } \lambda > 0,\\
        x^\lambda-1 \le 0 & \text{if } \lambda < 0.\\
    \end{cases}
\end{equation}
When $\lambda=0$, $\ln x\ge0$ for $x\ge1$. Hence $\psi(\lambda,x)\ge0$ for all $\lambda$ whenever $x\ge1$.
Similarly, for $0<x<1$, we have:
\begin{equation}
    \begin{cases}
        x^\lambda-1 < 0 & \text{if } \lambda > 0,\\
        x^\lambda-1 > 0 & \text{if } \lambda < 0.\\
    \end{cases}
\end{equation}
When $\lambda=0$, $\ln x<0$ for $0<x<1$. Hence $\psi(\lambda,x)<0$ for all $\lambda$ whenever $0<x<1$.
    
\item 
The second order partial derivative of $\psi$ with respect to $x$ is:
\begin{equation}
    \frac{\partial^2\psi(\lambda,x)}{\partial x^2}=
    \begin{cases}
        (\lambda-1)x^{\lambda-2} & \text{if } \lambda \neq 0,\\
        -1/x^2 & \text{if } \lambda = 0.\\
    \end{cases}
\end{equation}
Therefore, $\frac{\partial^2\psi(\lambda,x)}{\partial x^2}>0$ when $\lambda>1$ and $\frac{\partial^2\psi(\lambda,x)}{\partial x^2}<0$ when $\lambda<1$.

\item 
It's clear that $\psi(\lambda,x)$ is continuous for $\lambda$ and $x$ except $\lambda=0$. We just need to prove it's continuous at $\lambda=0$. By L'Hopital's rule, we have:
\begin{equation}
    \lim_{\lambda\to0}\frac{x^\lambda-1}{\lambda} = \lim_{\lambda\to0}\frac{x^\lambda\ln x}{1} = \ln x.
\end{equation}

\item 
We prove this by induction. Let $k=1$, then for $\lambda\neq0$:
\begin{equation}
    \psi^{(1)}(\lambda,x)
    = \frac{x^\lambda\lambda\ln x-(x^\lambda-1)}{\lambda^2}
    = \frac{x^\lambda\ln x-\psi^{(0)}(\lambda,x)}{\lambda}.
\end{equation}
Assume that this hold for $k=n$ where $n\ge1$, then for $k=n+1$ and $\lambda\neq0$:
\begin{align}
    \psi^{(n+1)}(\lambda,x)
    & = \frac{\partial}{\partial\lambda}\frac{x^\lambda(\ln x)^n-n\psi^{(n-1)}(\lambda,x)}{\lambda} \\
    & = \frac{[x^\lambda(\ln x)^{n+1}-n\psi^{(n)}(\lambda,x)]\lambda - [x^\lambda(\ln x)^n-n\psi^{(n-1)}(\lambda,x)]}{\lambda^2} \\
    & = \frac{x^\lambda(\ln x)^{n+1}-(n+1)\psi^{(n)}(\lambda,x)}{\lambda}.
\end{align}
For $\lambda=0$, recall the Taylor expansion of $x^\lambda$ at $\lambda=0$ is:
\begin{equation}
    x^\lambda = 1 + \lambda\ln x + \frac{(\lambda\ln x)^2}{2!} + \frac{(\lambda\ln x)^3}{3!} + \cdots
    = \sum_{m=0}^\infty \frac{(\lambda\ln x)^m}{m!}.
\end{equation}
Thus,
\begin{equation}
    \psi(\lambda,x)
    = \frac{x^\lambda - 1}{\lambda}
    = \ln x + \frac{\lambda(\ln x)^2}{2!} + \frac{\lambda^2(\ln x)^3}{3!} + \cdots
    = \sum_{m=1}^\infty \frac{\lambda^{m-1}(\ln x)^m}{m!}.
\end{equation}
Therefore, the $k$-th order partial derivative of $\psi$ with respect to $\lambda$ is:
\begin{equation}
    \frac{\partial^k\psi(\lambda,x)}{\partial\lambda^k}
    = \sum_{m=k+1}^\infty \frac{(m-1)!\lambda^{m-k-1}(\ln x)^m}{m!}
    = \sum_{m=k+1}^\infty \frac{\lambda^{m-k-1}(\ln x)^m}{m}.
\end{equation}
Finally, at $\lambda=0$, we have:
\begin{equation}
    \frac{\partial^k\psi(\lambda,x)}{\partial\lambda^k}\Biggr|_{\lambda=0} 
    = \frac{(\ln x)^{k+1}}{k+1}.
\end{equation}
This completes the induction.

\item 
The partial derivative of $\psi$ with respect to $x$ is
\begin{equation}
    \frac{\partial\psi(\lambda,x)}{\partial x}=
    \begin{cases}
        x^{\lambda-1} & \text{if } \lambda \neq 0,\\
        1/x & \text{if } \lambda = 0.\\
    \end{cases}
\end{equation}
so $\frac{\partial\psi(\lambda,x)}{\partial x}>0$. Therefore, $\psi$ is increasing in $x$.

The partial derivative of $\psi$ with respect to $\lambda$ is
\begin{equation}
    \frac{\partial\psi(\lambda,x)}{\partial\lambda}=
    \begin{cases}
        \frac{x^\lambda(\ln x^\lambda-1)+1}{\lambda^2} & \text{if } \lambda \neq 0,\\
        (\ln x)^2/2 & \text{if } \lambda = 0.\\
    \end{cases}
\end{equation}
Let $y=x^\lambda > 0$ and $f_1(y)=y(\ln y-1)+1$, we have $f_1'(y)=\ln y$, $f_1''(y)=1/y>0$. Thus $f_1(y)$ has the unique minimum at $y=1$ and $f_1(y)>f_1(1)=0$. Thus $\frac{\partial\psi(\lambda,x)}{\partial\lambda}>0$. Therefore, $\psi$ is increasing in $\lambda$.

\item 
The second order partial derivative of $\psi$ with respect to $\lambda$ is
\begin{equation}
    \frac{\partial^2\psi(\lambda,x)}{\partial\lambda^2}=
    \begin{cases}
        \frac{x^\lambda[(\ln x^\lambda)^2-2\ln x^\lambda + 2]-2}{\lambda^3} & \text{if } \lambda \neq 0,\\
        (\ln x)^3/3 & \text{if } \lambda = 0.\\
    \end{cases}
\end{equation}
Let $y=x^\lambda > 0$ and $f_2(y)=y[(\ln y)^2-2\ln y+2]-2$, we have $f_2'(y)=(\ln y)^2>0$ and $f_2(1)=0$. Thus $f_2(y)>0$ when $y>1$ and $f_2(y)<0$ when $y<1$ since $f_2(y)$ is increasing in $y$.

The relationship between $x,\lambda$ and $y,f_2(y)$ are as follows
\begin{equation}
    \begin{aligned}
        \left.
        \begin{aligned}
            & x>1, \lambda>0 & \Rightarrow y > 1, f_2(y)>0 \\
            & x>1, \lambda<0 & \Rightarrow y < 1, f_2(y)<0
        \end{aligned}
        \right\} \Rightarrow f_2(y)/\lambda^3>0 \\ 
        \left.
        \begin{aligned}
            & 0<x<1, \lambda<0 & \Rightarrow y > 1, f_2(y)>0 \\
            & 0<x<1, \lambda>0 & \Rightarrow y < 1, f_2(y)<0
        \end{aligned}
        \right\} \Rightarrow f_2(y)/\lambda^3<0 \\ 
    \end{aligned}
\end{equation}
Therefore, $\frac{\partial^2\psi(\lambda,x)}{\partial\lambda^2}>0$ when $x>1$ and $\frac{\partial^2\psi(\lambda,x)}{\partial\lambda^2}<0$ when $0<x<1$.
\end{enumerate}

\section{Log-domain Computation}
\label{sec:log-domain}
Log-domain computation refers to performing numerical operations in the logarithmic space rather than on raw values. This technique is particularly effective at avoiding numerical overflow and underflow, especially when working with exponential functions, as in the case of power transforms. Central to this approach is the \emph{Log-Sum-Exp} (LSE) function:
\begin{equation}
\text{LSE}(x_1,\dots,x_n) = \ln\sum_{i=1}^n \exp(x_i) = \ln\sum_{i=1}^n \exp(x_i-c)+c
\end{equation}
where $c = \max_i x_i$. This ensures numerically stable computation by shifting values before exponentiation.

Using the LSE operator, the logarithmic mean is computed as
\begin{equation}
\ln\mu = \ln \left( \frac{1}{n}\sum_{i=1}^n x_i \right)
= \text{LSE}(\ln x_1,\dots,\ln x_n)-\ln n
\label{eq:log_mean}
\end{equation}
Similarly, the logarithmic variance is expressed as
\begin{equation}
	\label{eq:log_var}
	\ln\sigma^2 = \ln\sum_{i=1}^n (x_i-\mu)^2-\ln n = \text{LSE}\left(2\ln(x_1-\mu),\dots,2\ln(x_n-\mu)\right)-\ln n
\end{equation}
The computation of $\ln(x_i - \mu)$ requires extra care. A stable way is to rewrite the difference using the LSE trick \citep{logvar}:
\begin{equation}
	\ln(x_i-\mu) = \ln\left(\exp(\ln x_i)+\exp(\ln\mu+\pi i)\right) = \text{LSE}(\ln x_i,\ln\mu+\pi i)
\end{equation}
where the $\pi i$ term is the imaginary part that handles sign differences for negative values.

\section{Numerically Stable Yeo-Johnson}
\label{sec:stable-yj}
To extend numerical stability to the Yeo-Johnson transformation, we must handle both positive and negative inputs. Unlike Box-Cox, the constant term cannot be eliminated when both positive and negative values are present. We therefore consider the following two cases separately:
\begin{enumerate}
    \item Data entirely positive (or entirely negative): Apply log-domain computation as in Box-Cox, removing the constant term and factoring out $\lambda$ (for positive) or $(\lambda-2)$ (for negative).
    \item Data contains both positive and negative values: Use the full piecewise definition of Yeo-Johnson (Equation~\eqref{eq:yeojohnson}) with log-domain computation, but do not remove the constant term. In practice, this case does not exhibit instability.
\end{enumerate}
As in the Box-Cox case, extreme values of $\lambda$ can be constrained during optimization. Here, we use $x=0$ as the reference point since $\psi_\text{YJ}(\lambda,0)=0$:
\begin{mini}|s|
	{\lambda}{\text{NLL}_{\text{YJ}}(\lambda, x)}
	{}{}
	\addConstraint{\lambda\le\psi^{-1}_{\text{YJ}}(x_{\text{max}},y_B)\ \ \ \text{if } x_{\text{max}}>0}
	\addConstraint{\lambda\ge\psi^{-1}_{\text{YJ}}(x_{\text{min}},-y_B)\ \text{if } x_{\text{min}}<0}.
	\label{yj-constrained}
\end{mini}
Here $\psi^{-1}_\text{YJ}$ is the inverse Yeo-Johnson transformation, expressed using the Lambert $W$ function:
\begin{equation}
\label{eq:inv-yeojohnson}
\psi^{-1}_{\text{YJ}}(x, y) =
\begin{cases}
    -\tfrac{1}{y}-\tfrac{1}{\ln(x+1)}W\left(-\tfrac{(x+1)^{-1/y}\ln(x+1)}{y}\right) & \text{if } x\ge0,\\
    2-\tfrac{1}{y}+\tfrac{1}{\ln(1-x)}W\left(\tfrac{(1-x)^{1/y}\ln(1-x)}{y}\right) & \text{if } x<0.\\
\end{cases}
\end{equation}
For overflow cases, we employ the $k=-1$ branch of the Lambert $W$ function. As for $x>0$ and $\lambda\gg 1$:
\begin{equation}
W\left(-\tfrac{(x+1)^{-1/y}\ln(x+1)}{y}\right)
=-(\lambda+\tfrac{1}{y})\ln(x+1)
\approx -\lambda\ln(x+1) \ll -1 .
\end{equation}
For $x<0$ and $\lambda\ll 1$:
\begin{equation}
W\left(\tfrac{(1-x)^{1/y}\ln(1-x)}{y}\right)
=(\lambda+\tfrac{1}{y}-2)\ln(1-x)
\approx (\lambda-2)\ln(1-x) \ll -1 .
\end{equation}
\section{Federated NLL Computation for Yeo-Johnson}
\label{sec:fed-nll-yj}
In the federated setting, the Yeo-Johnson transformation requires additional care due to its piecewise definition for positive and negative inputs. The main challenges are: (1) clients need to use different formulas to achieve numerical stability based on their local data (all-positive, all-negative, or mixed); (2) the server must correctly aggregate statistics from these heterogeneous clients.

First, instead of reporting only the total number of samples, each client separately transmits the counts of positive and negative values. This enables the server to identify whether the client's dataset is all-positive, all-negative, or mixed, and to aggregate contributions accordingly.

Second, depending on the case, clients apply different formulas for the transformed data. In the all-positive (or all-negative) case, constant term can be safely omitted. For mixed data, however, the full piecewise definition of the  transform function must be used to preserve correctness of the variance.

Finally, the server aggregates the statistics and computes the variance of the transformed data. It then applies a correction step to adjust the mean, compensating for constant term that clients may have omitted. The complete procedure is summarized in Algorithm~\ref{alg:fed-nll-yj}.
\begin{algorithm}
\caption{Federated NLL Computation (Yeo-Johnson)}
\label{alg:fed-nll-yj}
\begin{algorithmic}[1]
\Statex \textbf{Client Part:}
\State \textbf{Input:} $\lambda$ and local data $x_i$ (size $n$, with positive size $n^+$ and negative size $n^-$)
\State $c \gets \sum_i \text{sgn}(x_i) \ln(|x_i|+1)$
\If{$n^-=0$} \Comment{All positive}
\State $y_i \gets
\begin{cases}
(x_i+1)^\lambda & \text{if } \lambda\neq0,\\
\ln(x_i+1) & \text{if } \lambda=0.
\end{cases}$
\ElsIf{$n^+=0$} \Comment{All negative}
\State $y_i \gets
\begin{cases}
(-x_i+1)^{2-\lambda} & \text{if } \lambda\neq2,\\
-\ln(-x_i+1) & \text{if } \lambda=2.
\end{cases}$
\Else \Comment{Mixed data}
\State $y_i \gets \psi_\text{YJ}(\lambda,x_i)$ (Equation~\eqref{eq:yeojohnson})
\EndIf
\State $\bar{y} \gets \tfrac{1}{n} \sum_i y_i$ and $s \gets \sum_i (y_i-\bar{y})^2$
\State \textbf{Send:} $(c, n^+, n^-, \bar{y}, s)$ to server
\setcounter{ALG@line}{0}

\Statex
\Statex \textbf{Server Part:}
\State \textbf{Input:} $\lambda$
\State Collect $(c^{(j)}, n^{+{(j)}}, n^{-{(j)}}, \bar{y}^{(j)}, s^{(j)})$ from clients
\If{$n^{-(j)}=0$} \Comment{All positive}
\State Enqueue $(n^{+(j)}, \bar{y}^{(j)}, s^{(j)})$ into $Q^+$
\ElsIf{$n^{+(j)}=0$}\Comment{All negative}
\State Enqueue $(n^{-(j)}, \bar{y}^{(j)}, s^{(j)})$ into $Q^-$
\Else \Comment{Mixed data}
\State Enqueue $(n^{+(j)}+n^{-(j)}, \bar{y}^{(j)}, s^{(j)})$ into $Q^\pm$
\EndIf
\State Compute $(N^+, \bar{Y}^+, S^+)$ from $Q^+$ using Algorithm~\ref{alg:var-agg}
\If{$\lambda\neq0$}
\State $\ln S^+ \gets \ln S^+ - 2\ln|\lambda|$
\EndIf
\State Compute $(N^-, \bar{Y}^-, S^-)$ from $Q^-$ using Algorithm~\ref{alg:var-agg}
\If{$\lambda\neq2$}
\State $\ln S^- \gets \ln S^- - 2\ln|2-\lambda|$
\EndIf
\State Compute $(N^\pm, \bar{Y}^\pm, S^\pm)$ from $Q^\pm$ using Algorithm~\ref{alg:var-agg}
\If{($N^->0$ or $N^\pm>0$) and $\lambda\neq0$} \Comment{Add constant term back for mixed data}
\State $\bar{Y}^+ \gets (\bar{Y}^+ - 1) / \lambda$
\ElsIf{($N^+>0$ or $N^\pm>0$) and $\lambda\neq2$}
\State $\bar{Y}^- \gets (\bar{Y}^- - 1) / (\lambda-2)$
\EndIf
\State Enqueue $(N^+, \bar{Y}^+, S^+)$, $(N^-, \bar{Y}^-, S^-)$, $(N^\pm, \bar{Y}^\pm, S^\pm)$ into $Q$
\State Compute $(N, \bar{Y}, S)$ from $Q$ using Algorithm~\ref{alg:var-agg}
\State $\ln\sigma^2_{\psi} \gets \ln S - \ln N$
\State $\text{NLL}_\text{YJ} \gets (1-\lambda)\sum_j c^{(j)} + \tfrac{N}{2}\ln\sigma^2_{\psi}$
\State \Return $\text{NLL}_\text{YJ}$
\end{algorithmic}
\end{algorithm}

\section{Supplementary Tables and Plots}
\label{sec:supp}
\begin{table}[ht]
\caption{Dataset statistics.}
\label{tab:train-data-stats}
\centering
\begin{tabular}{cccc}
\toprule
Datasets & \# Row & \# Col \\
\midrule
Adult & 33K & 14 \\
Bank & 45K & 16 \\
Credit & 30K & 24 \\
\bottomrule
\end{tabular}
\end{table}
\begin{figure}[ht]
\centering
\includegraphics[width=0.33\columnwidth]{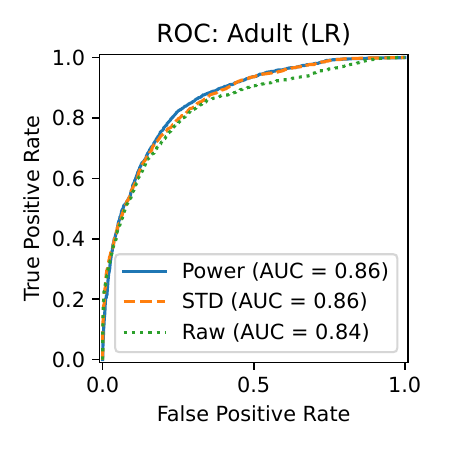}%
\includegraphics[width=0.33\columnwidth]{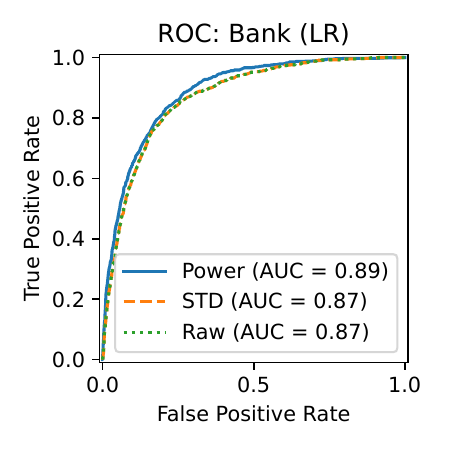}%
\includegraphics[width=0.33\columnwidth]{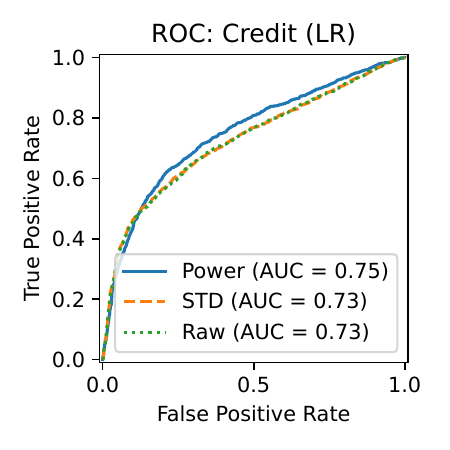}\\
\includegraphics[width=0.33\columnwidth]{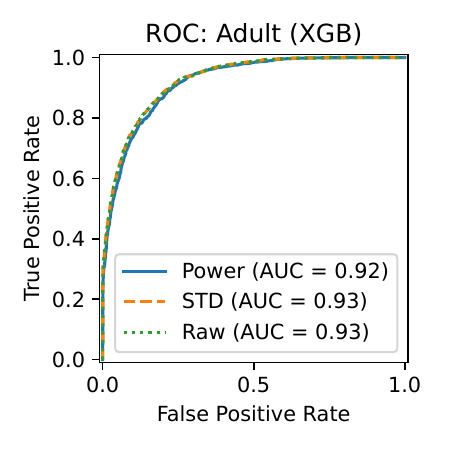}%
\includegraphics[width=0.33\columnwidth]{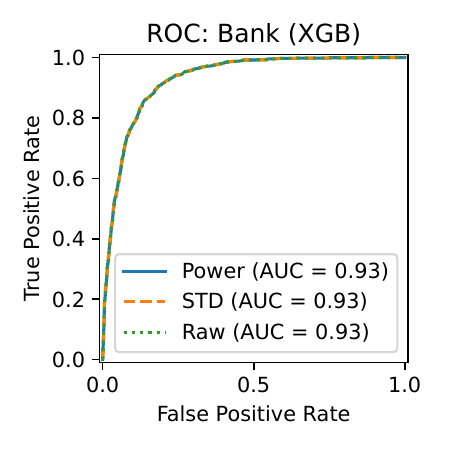}%
\includegraphics[width=0.33\columnwidth]{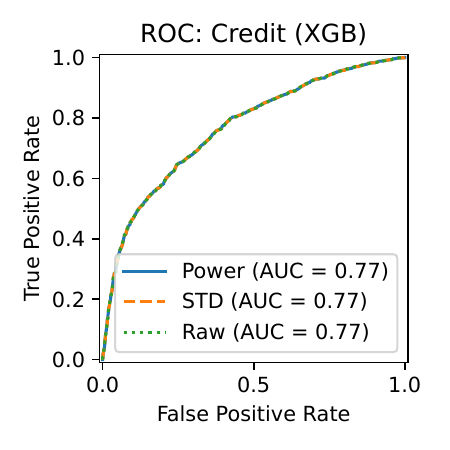}%
\caption{ROC curves after applying different transforms (LR and XGB).}%
\label{fig:effect-roc-LR-XGB}%
\end{figure}
\begin{figure}[ht]
\centering
\includegraphics[width=0.33\columnwidth]{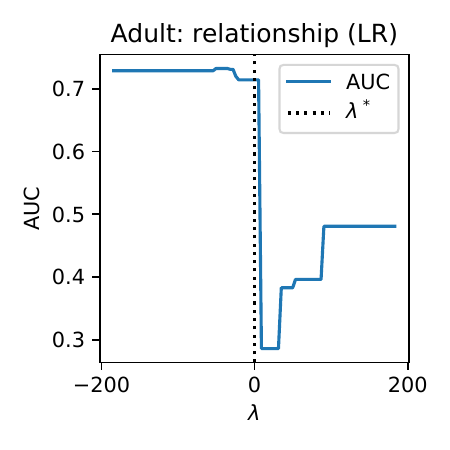}%
\includegraphics[width=0.33\columnwidth]{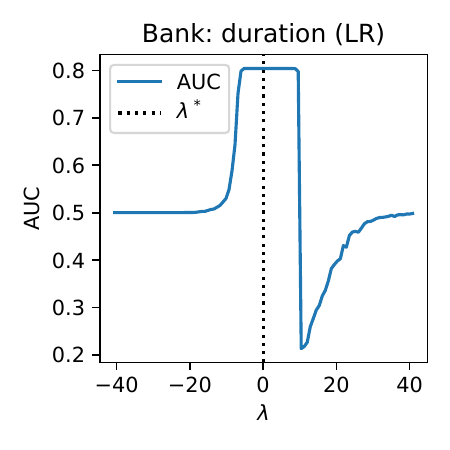}%
\includegraphics[width=0.33\columnwidth]{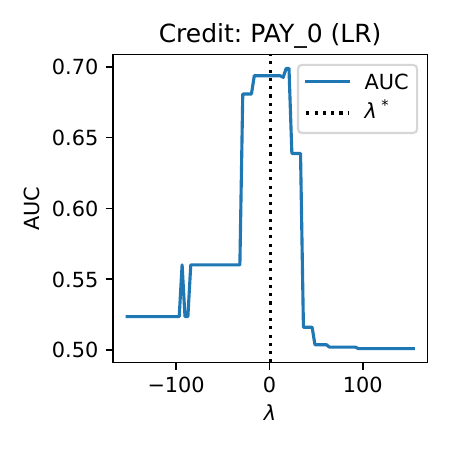}\\
\includegraphics[width=0.33\columnwidth]{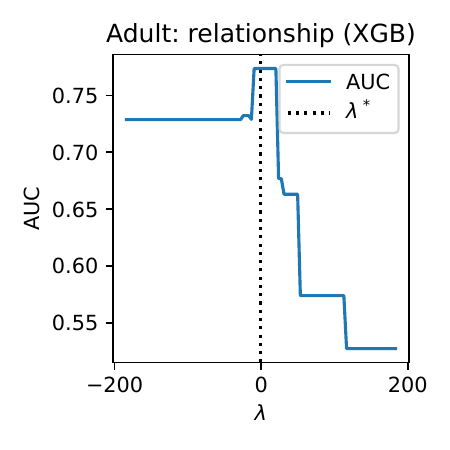}%
\includegraphics[width=0.33\columnwidth]{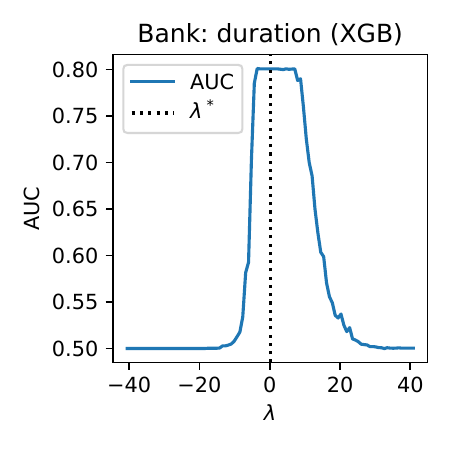}%
\includegraphics[width=0.33\columnwidth]{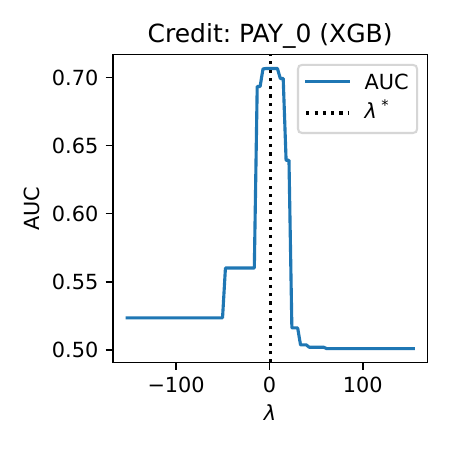}%
\caption{Effect of varying $\lambda$ on AUC scores (LR and XGB).}%
\label{fig:deviate-auc-LR-XGB}%
\end{figure}
\begin{figure}[ht]
\centering
\includegraphics[width=0.33\columnwidth]{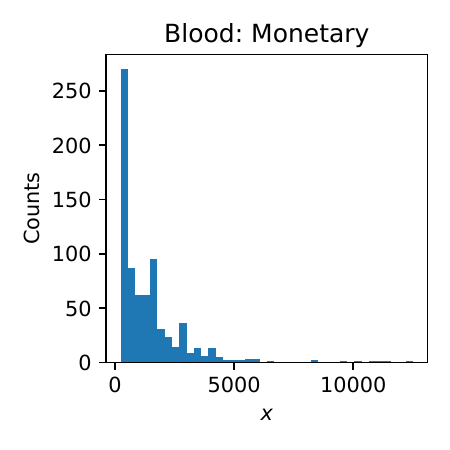}%
\includegraphics[width=0.33\columnwidth]{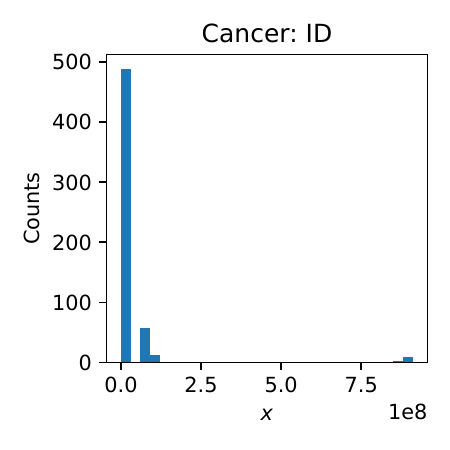}%
\includegraphics[width=0.33\columnwidth]{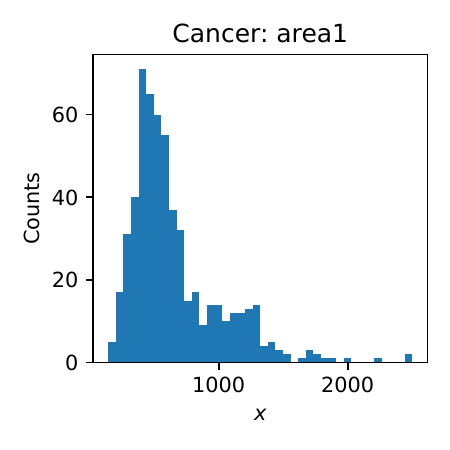}\\
\includegraphics[width=0.33\columnwidth]{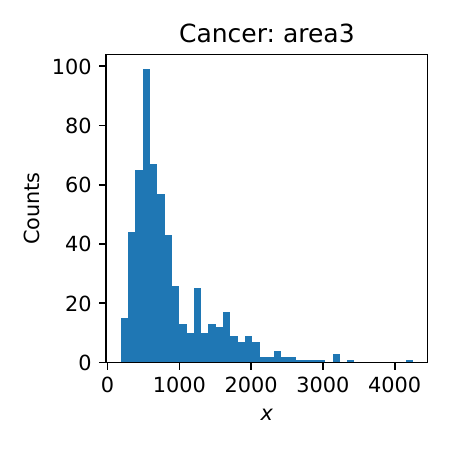}%
\includegraphics[width=0.33\columnwidth]{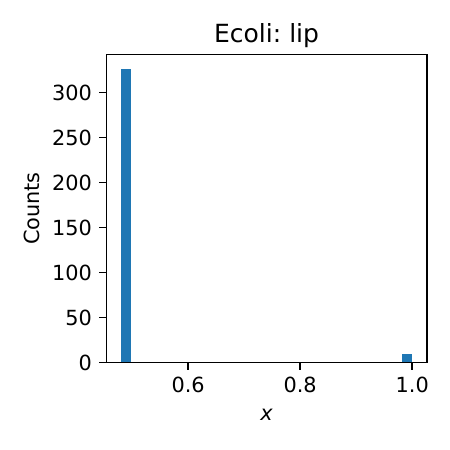}%
\includegraphics[width=0.33\columnwidth]{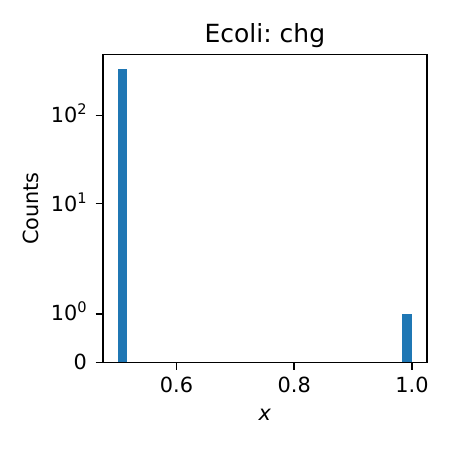}\\
\includegraphics[width=0.33\columnwidth]{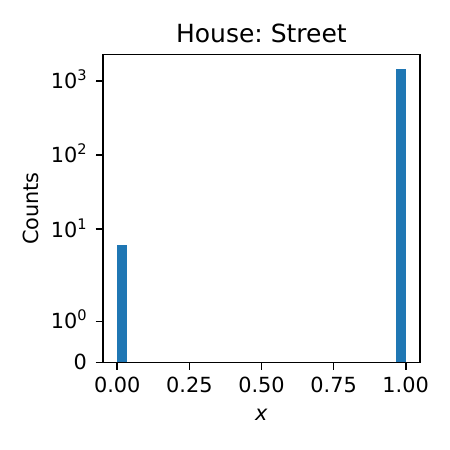}%
\includegraphics[width=0.33\columnwidth]{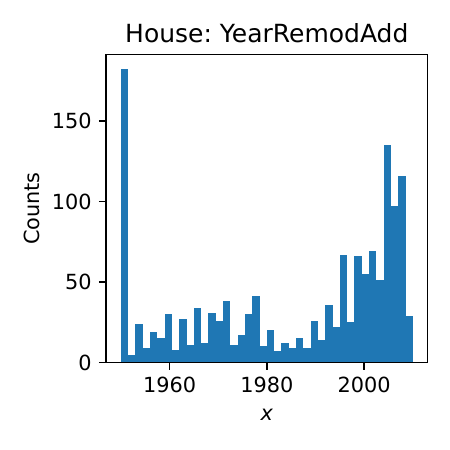}%
\includegraphics[width=0.33\columnwidth]{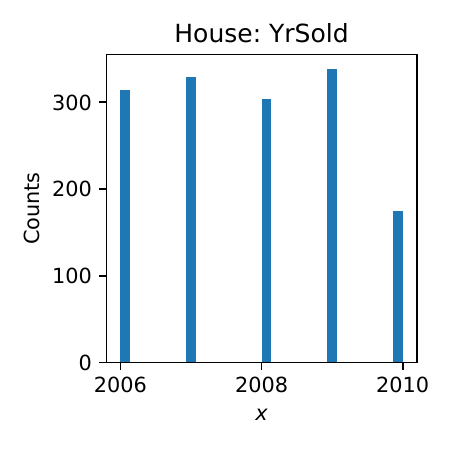}%
\caption{Histogram of features that have numerical issues.}%
\label{fig:feature-hist}%
\end{figure}
\begin{figure}[ht]
\centering
\includegraphics[width=\columnwidth]{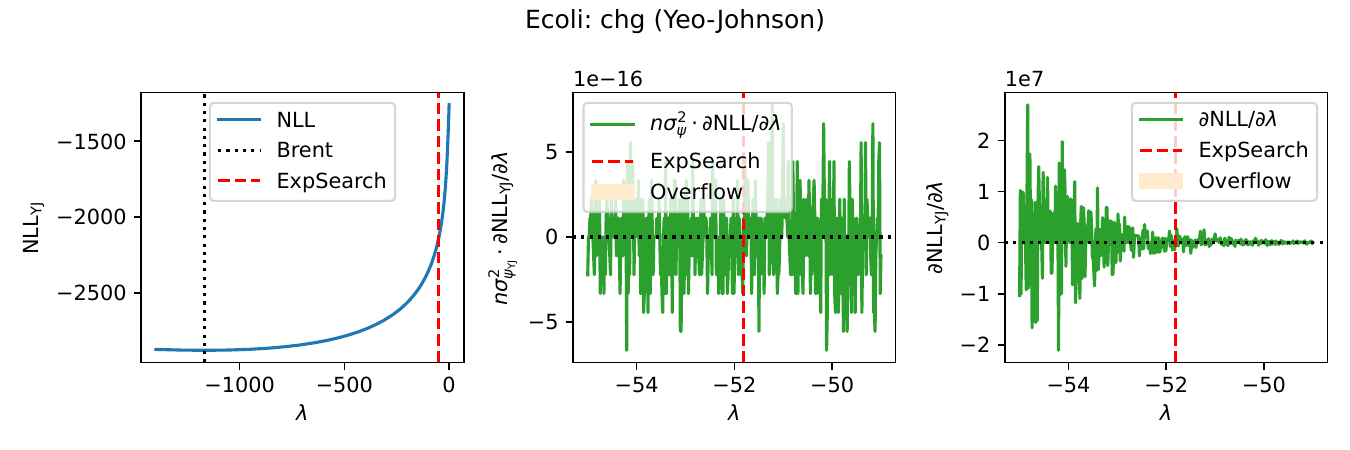}\\
\includegraphics[width=\columnwidth]{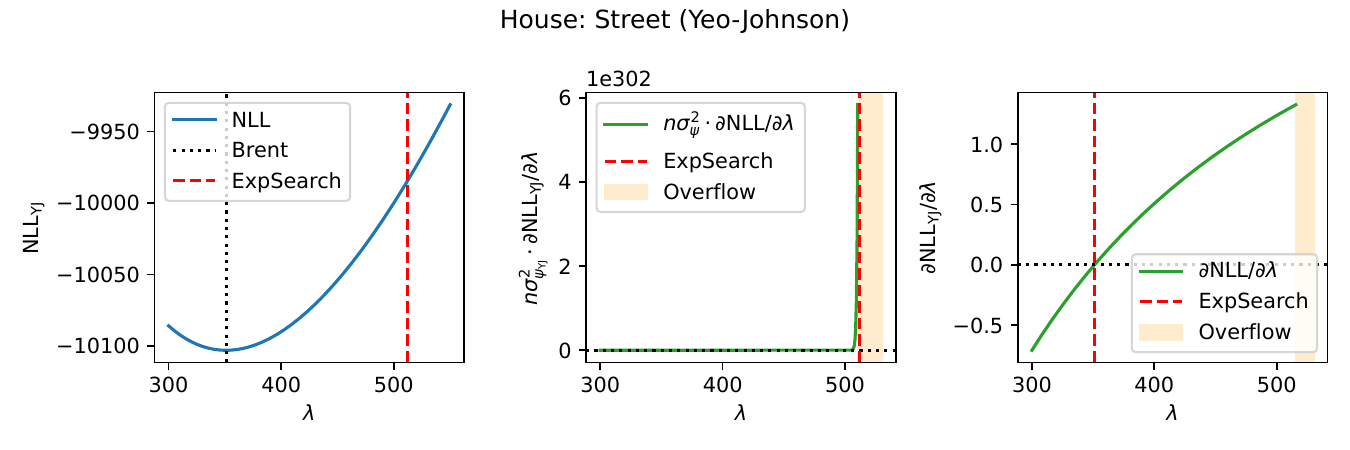}\\
\includegraphics[width=0.99\columnwidth]{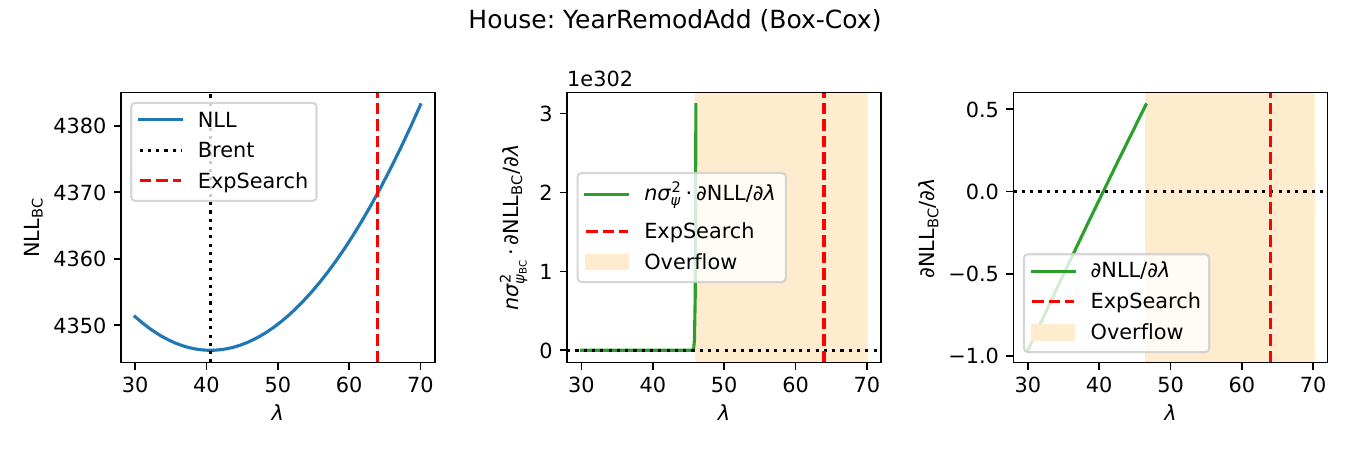}%
\caption{Comparison of Brent's method and ExpSearch.}%
\label{fig:brent-expsearch-appendix-1}%
\end{figure}
\begin{figure}[ht]
\ContinuedFloat
\centering
\includegraphics[width=\columnwidth]{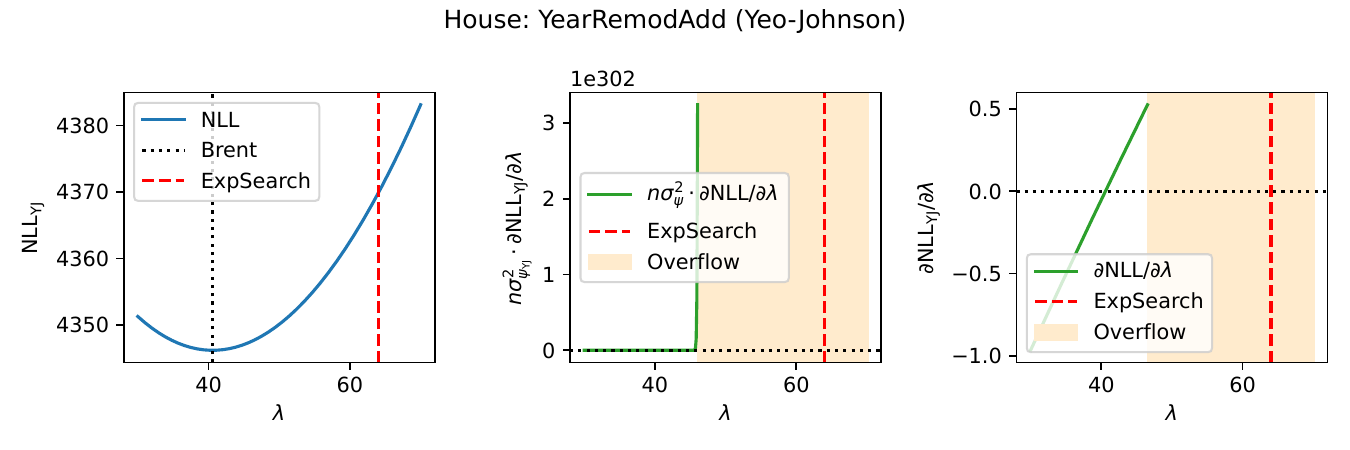}\\
\includegraphics[width=\columnwidth]{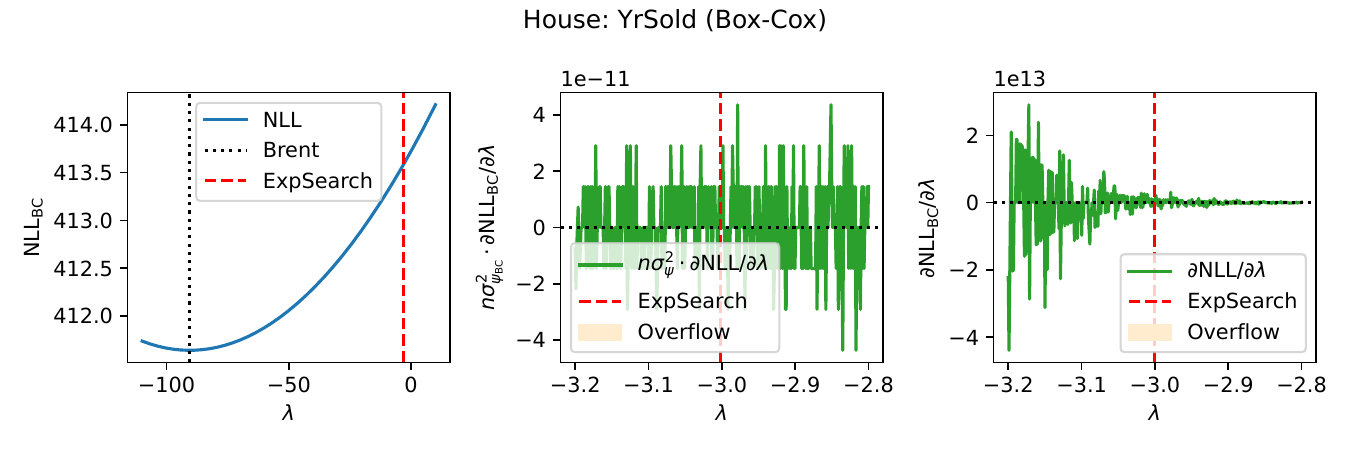}\\
\includegraphics[width=\columnwidth]{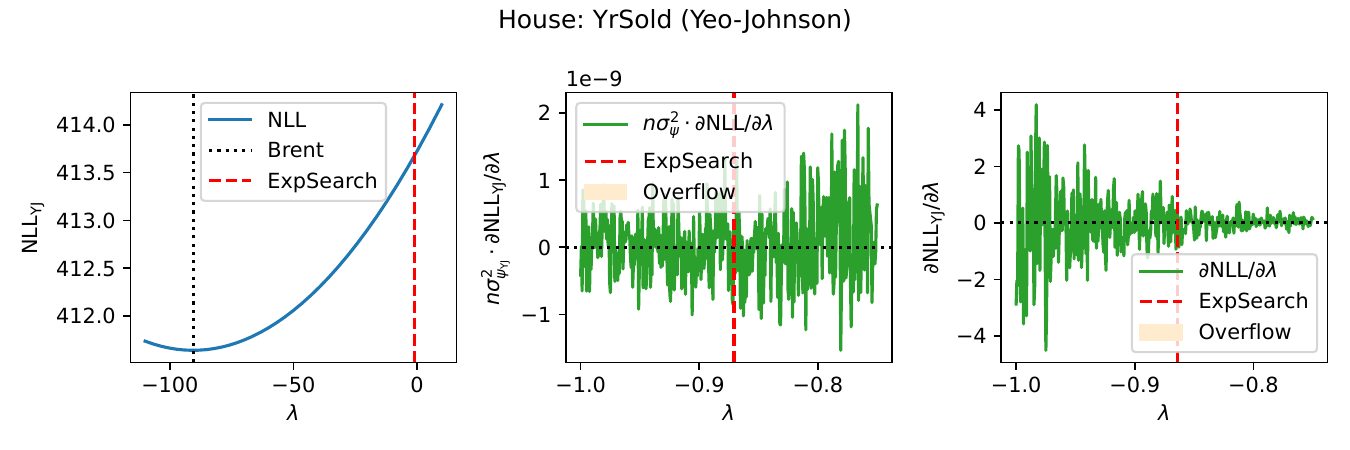}%
\caption{Comparison of Brent's method and ExpSearch.}%
\label{fig:brent-expsearch-appendix-2}%
\end{figure}
\begin{figure}[ht]
\centering
\includegraphics[width=0.33\columnwidth]{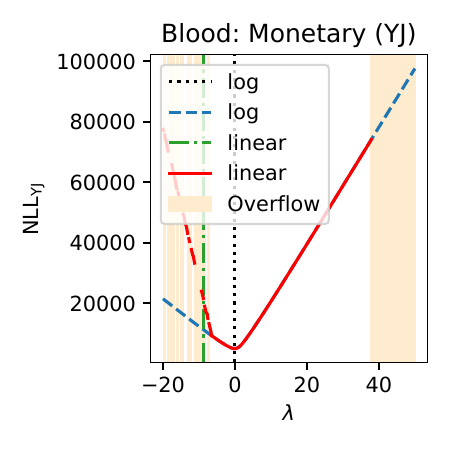}%
\includegraphics[width=0.33\columnwidth]{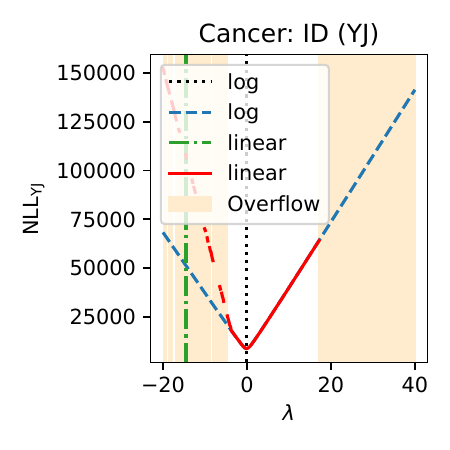}%
\includegraphics[width=0.33\columnwidth]{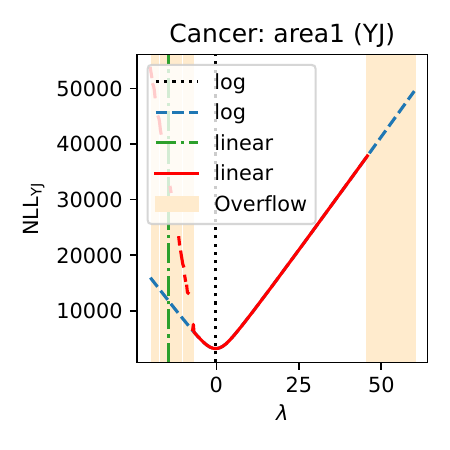}\\
\includegraphics[width=0.33\columnwidth]{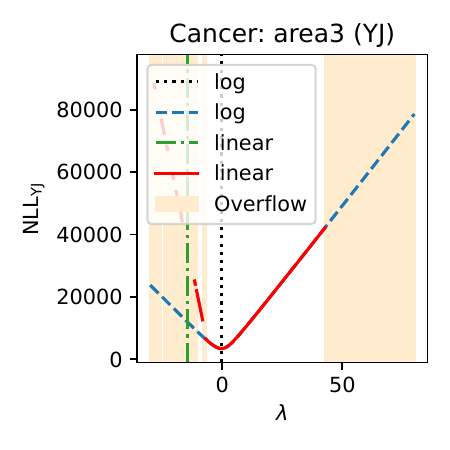}%
\includegraphics[width=0.33\columnwidth]{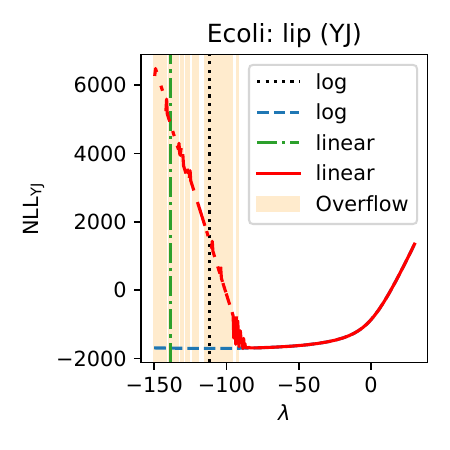}%
\includegraphics[width=0.33\columnwidth]{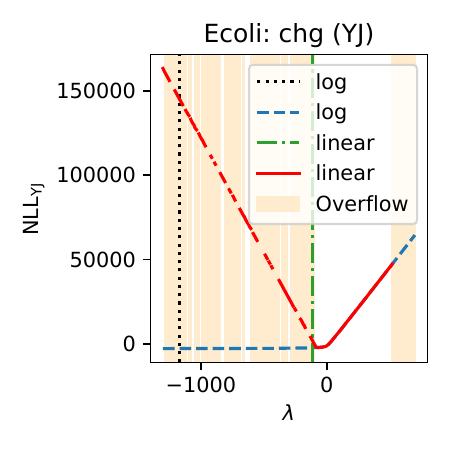}\\
\includegraphics[width=0.33\columnwidth]{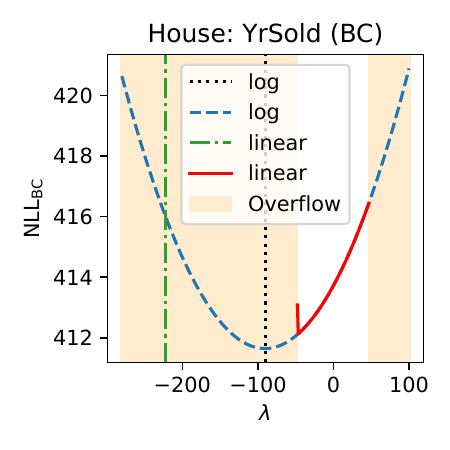}%
\includegraphics[width=0.33\columnwidth]{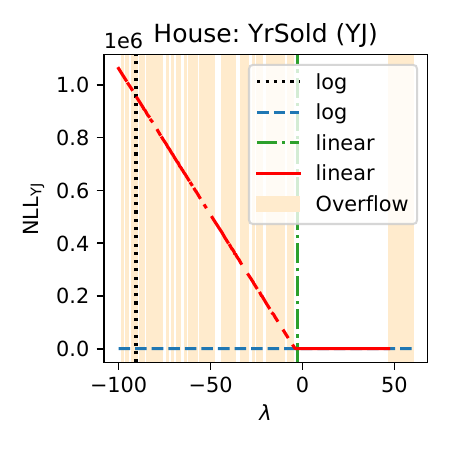}%
\caption{Comparison of log and linear domain computation.}%
\label{fig:log-linear}%
\end{figure}
\begin{figure}[ht]
\centering
\includegraphics[width=0.33\columnwidth]{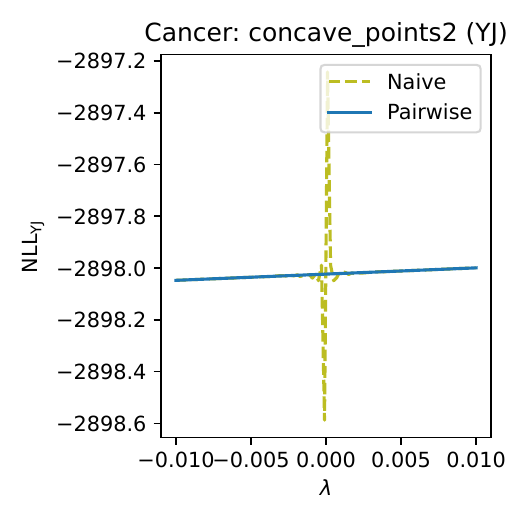}%
\includegraphics[width=0.33\columnwidth]{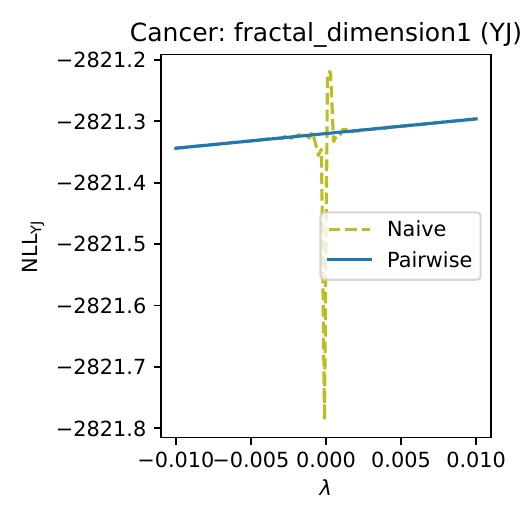}%
\includegraphics[width=0.33\columnwidth]{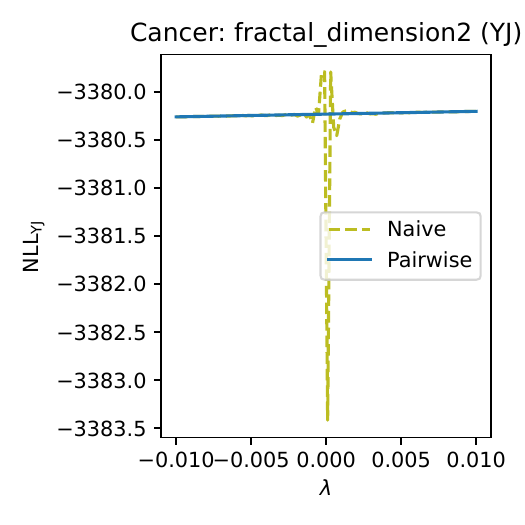}\\
\includegraphics[width=0.33\columnwidth]{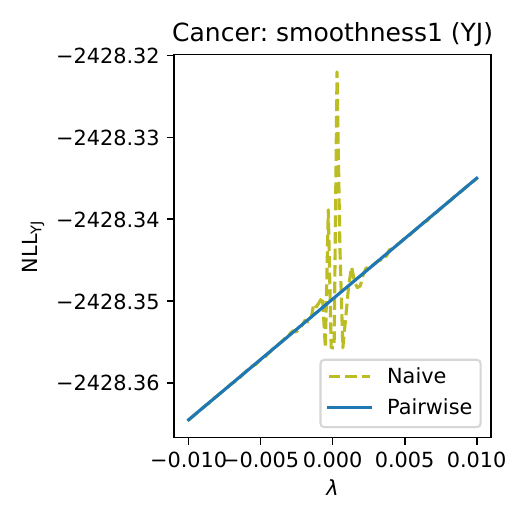}%
\includegraphics[width=0.33\columnwidth]{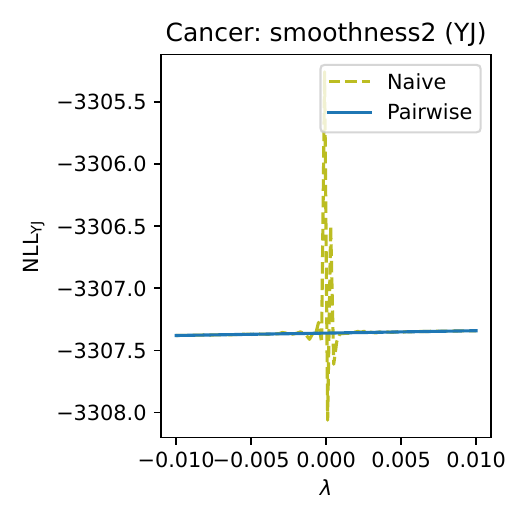}%
\includegraphics[width=0.33\columnwidth]{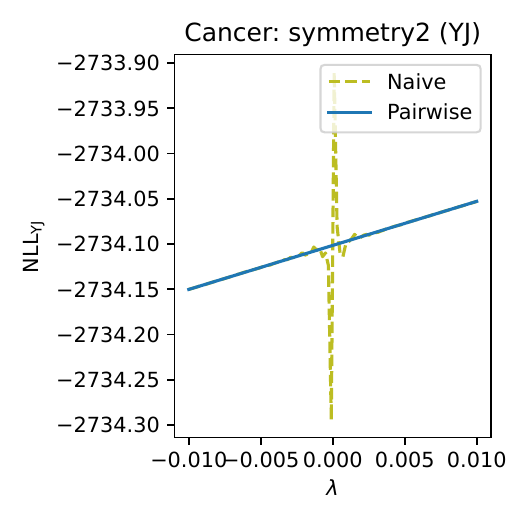}\\
\includegraphics[width=0.33\columnwidth]{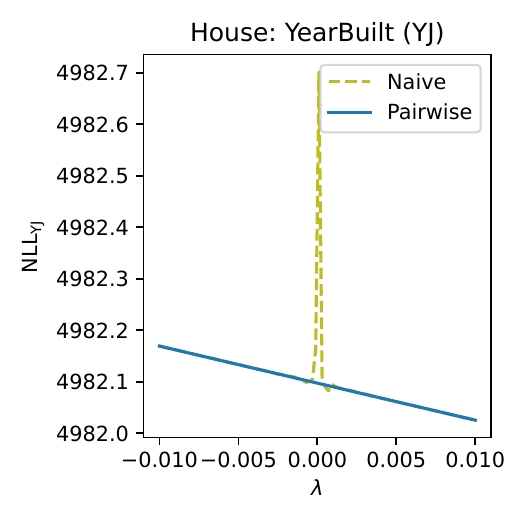}%
\includegraphics[width=0.33\columnwidth]{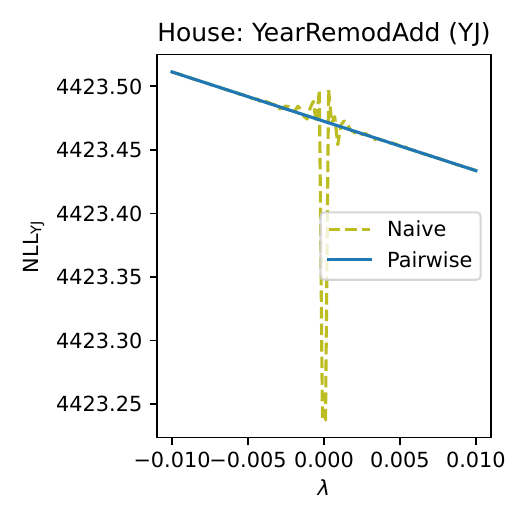}%
\includegraphics[width=0.33\columnwidth]{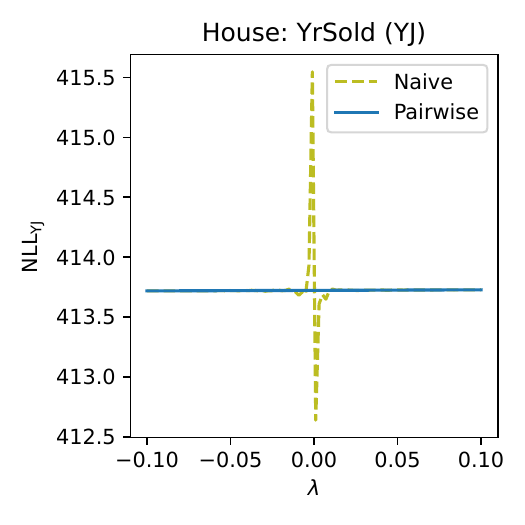}%
\caption{Comparison of pairwise and naive variance aggregation.}%
\label{fig:pairwise-naive-appendix}%
\end{figure}
\begin{table}[ht]
\caption{Runtime comparison between naive and our method for Box-Cox NLL computation.}
\label{tab:time-compare}
\centering
\begin{tabular}{ccc}
\toprule
Data size & Time (Naive) & Time (Ours) \\
\midrule
1K & \SI{13.10}{\micro\second} $\pm$ \SI{391.62}{\nano\second} & \SI{273.88}{\micro\second} $\pm$ \SI{819.93}{\nano\second} \\
10K & \SI{51.70}{\micro\second} $\pm$ \SI{2.54}{\micro\second} & \SI{1.70}{\milli\second} $\pm$ \SI{13.29}{\micro\second} \\
100K & \SI{604.00}{\micro\second} $\pm$ \SI{10.06}{\micro\second} & \SI{18.13}{\milli\second} $\pm$ \SI{183.97}{\micro\second} \\
1M & \SI{6.57}{\milli\second} $\pm$ \SI{60.04}{\micro\second} & \SI{194.75}{\milli\second} $\pm$ \SI{1.31}{\milli\second} \\
\bottomrule
\end{tabular}
\end{table}
\begin{figure}[ht]
\centering
\includegraphics[width=0.33\columnwidth]{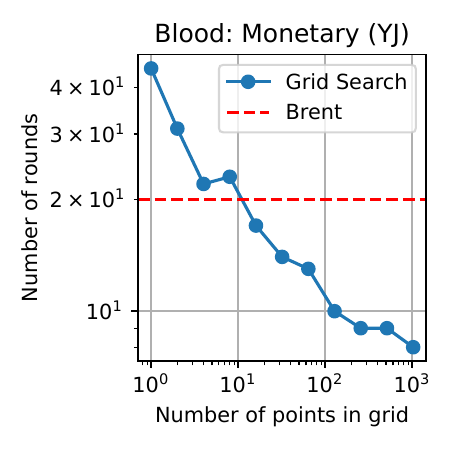}%
\includegraphics[width=0.33\columnwidth]{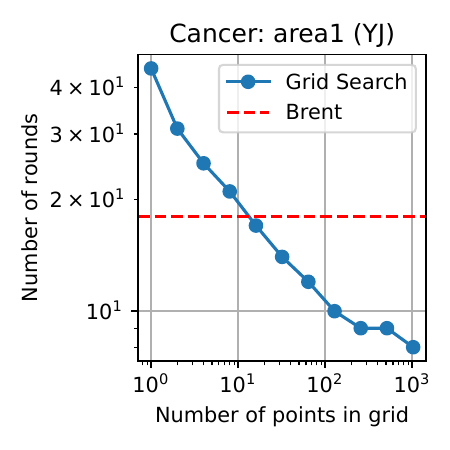}%
\includegraphics[width=0.33\columnwidth]{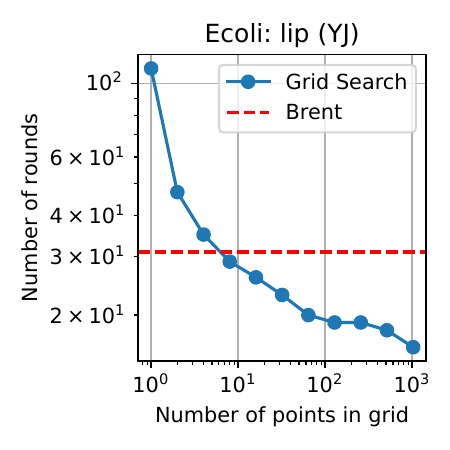}\\
\includegraphics[width=0.33\columnwidth]{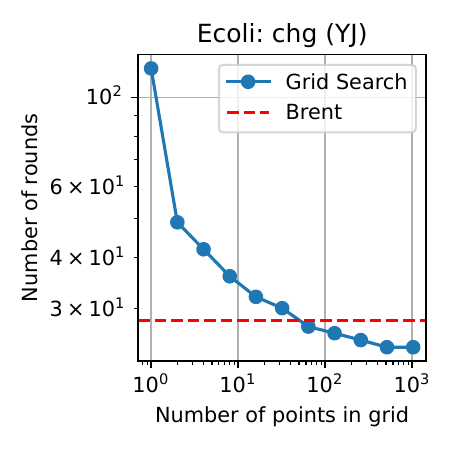}%
\includegraphics[width=0.33\columnwidth]{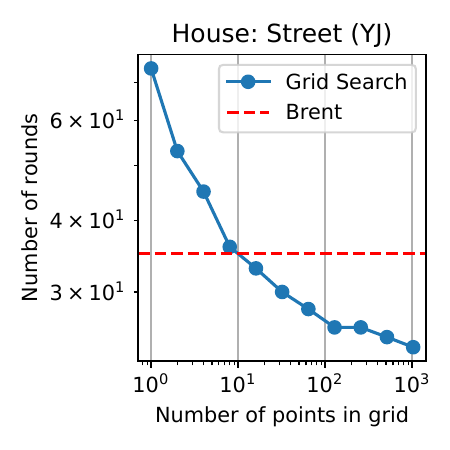}%
\includegraphics[width=0.33\columnwidth]{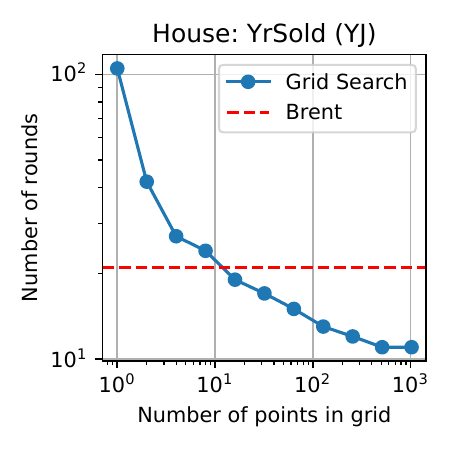}%
\caption{Number of communication rounds vs. grid size.}%
\label{fig:comm-round-grid}%
\end{figure}

\end{document}